# Quantum-Classical Physics-Informed Kolmogorov-Arnold Networks for Solving Fuzzy Differential Equations

**Xiang Rao[1,2,3,4,5]*, Yuxuan Shen[2]***

1. School of Petroleum Engineering, Yangtze University, Wuhan 430100, China
2. College of Future Technology, Yangtze University, Wuhan 430100, China
3. School of Computer Science, Yangtze University, Jingzhou 434023, China
4. State Key Laboratory of Low Carbon Catalysis and Carbon Dioxide Utilization (Yangtze University), Wuhan 430100, China
5. Western Research Institute, Yangtze University, Karamay 834000, China

*Corresponding author: Prof. Xiang Rao (raoxiang0103@163.com, raoxiang@yangtzeu.edu.cn), Ms. Yuxuan Shen (syx709@163.com)

Abstract: In this study, we propose a quantum-classical physics-informed Kolmogorov-Arnold network (QCPIKAN) dedicated to the solution of fuzzy differential equations. The network takes the spatiotemporal coordinates and membership level as joint inputs and employs ChebyKAN modules and a parameterized quantum circuit to construct a hybrid function approximator. It simultaneously approximates the lower and upper endpoint functions associated with the $\alpha$-cuts and incorporates the governing equations, initial-boundary conditions, and fuzzy-structural constraints into the training objective. Theoretically, a unified error-analysis framework is established for QCPIKAN and PIKAN, in which the endpoint-solution error is decomposed into approximation, sampling, optimization, and fuzzy-structure constraint errors. Under the assumptions of well-posedness and residual stability, it is proved that QCPIKAN has a smaller a priori error bound when the representational gain introduced by quantum entanglement features exceeds the additional computational error. Numerical experiments are conducted for elliptic, parabolic, and hyperbolic equations in an ideal quantum-simulation environment. The results show that QCPIKAN captures the overall contraction of the solution interval as $\alpha$ increases. At most tested membership levels, the mean relative $L_2$ error of PIKAN is approximately 1.1-2.7 times that of QCPIKAN. In the fuzzy convection example, the mean wavefront-position error of PIKAN is approximately 1.77 times that of QCPIKAN. Nevertheless, both models still exhibit local fuzzy-structure violations near boundaries, in high-gradient regions, and around the wavefront. These results indicate that QCPIKAN provides a quantum-classical hybrid physics-informed computational framework with comparatively high predictive accuracy for solving fuzzy partial differential equations represented by $\alpha$-cuts.

## 1 Introduction

In practical engineering and scientific computing, information such as material parameters, external loads, initial states, and boundary conditions may be affected by measurement errors, insufficient samples, and epistemic limitations, making it difficult to describe using precise values or reliable probability distributions. Fuzzy sets use membership functions to represent the degree to which an element belongs to a set, thereby providing a mathematical description of such imprecise information [1]. In partial differential equations, when data associated with equation parameters or boundary conditions are fuzzy, the corresponding uncertainty can be characterized using fuzzy numbers or fuzzy-valued functions. Buckley and Feuring investigated such problems and discussed a class of elementary fuzzy partial differential equations (FPDEs) and their fuzzy solutions [2]. Subsequent FPDE studies have addressed different types of equations and sources of uncertainty, including heat-conduction equations with fuzzy initial data [3], Burgers equations with fuzzy parameters, fuzzy initial-boundary data, and fuzzy source terms [4], and structural-mechanics models with uncertain parameters [5].

Numerical solutions of FPDEs must not only approximate the governing equations but also appropriately represent the uncertainty range of the solution induced by fuzzy inputs. Therefore, developing FPDE solvers that balance computational accuracy, solution efficiency, and the ability to represent fuzzy information is of considerable theoretical significance and engineering value.

The $\alpha$-cut represents a fuzzy number through a nested family of level-set intervals [6]. At a prescribed membership level $\alpha$, a fuzzy parameter or fuzzy datum can be expressed as a closed interval $\left[\underline{z}(\alpha), \bar{z}(\alpha)\right]$, allowing the corresponding interval problem to be examined across different membership levels. In the lower-upper parametric representation, the lower-endpoint function is nondecreasing with respect to $\alpha$, whereas the upper-endpoint function is nonincreasing; equivalently, an interval at a higher membership level is contained in that at a lower level [7]. Hence, interval solutions computed at different $\alpha$ levels cannot be treated independently if they are to define a valid fuzzy-valued solution. At each fixed spatiotemporal location, the lower endpoint must not exceed the upper endpoint, and the resulting family of solution intervals must satisfy inter-level nesting. For the explicit computation of $\alpha$-cut endpoints, Buckley and Qu examined output $\alpha$-cuts obtained by mapping fuzzy inputs through a function and identified the conditions under which direct interval arithmetic yields exact output $\alpha$-cuts [8]. For general nonmonotone continuous functions, Scheerlinck et al. formulated endpoint determination at different $\alpha$ levels as a set of interrelated optimization problems and compared several endpoint-optimization algorithms [9]. For the numerical solution of FPDEs, Allahviranloo used finite-difference methods to approximate fuzzy reachable sets and demonstrated the approach using linear and nonlinear FPDE examples [10].

In the solution of deterministic differential equations, directly representing an unknown solution with a trainable function approximator has gradually emerged as a computational paradigm distinct from conventional discretization-based methods. Dissanayake and Phan-Thien represented approximate solutions of partial differential equations using neural-network universal approximators and combined them with a collocation method to transform equation solving into an unconstrained minimization problem [11]. Lagaris et al. constructed a trial solution from a fixed component satisfying the initial or boundary conditions and a feedforward neural-network component with adjustable parameters, and trained the network so that the trial solution satisfied the governing equation [12]. With the development of neural-network function approximation for differential equations, Raissi et al. proposed the physics-informed neural network (PINN) framework for forward and inverse problems involving nonlinear partial differential equations. By incorporating the physical laws described by the governing equations into network training, the neural-network approximation is subjected to the corresponding physical constraints [13]. From the broader perspective of physics-informed machine learning, Karniadakis et al. subsequently summarized the principal approaches for integrating data with mathematical models and their applications to forward solution, inverse analysis, and model discovery [14]. PINNs have been applied to physical modeling and parameter inversion in numerous fields, including inverse scattering and dielectric-parameter inversion in nano-optics and metamaterials [15], reconstruction of cardiac electrical activation propagation [16], parameter inversion and surrogate modeling in solid mechanics [17], identification of biological dynamical systems [18], inversion of frictional-contact temperature fields [19], and state and parameter estimation for subsurface flow [20]. Related studies have also improved PINN training through loss functions, sampling strategies, and representations of the governing physics, including uncertainty-weighted losses for manufacturing-system state prediction [21], residual-adaptive sampling for two-phase flow in porous media [22], weak-form physical constraints for subsurface single- and two-phase flow [23], and mixed pressure-velocity formulations for flow in heterogeneous porous media [24].

Fuzzy logic has also been incorporated into physics-informed learning models. Zarzycki and Ławryńczuk used a fuzzy data-fusion module to combine the outputs of a first-principles model and a GRU network and applied the resulting approach to process modeling and model predictive control [25]. Deng et al. proposed a physics-informed spatial fuzzy system that represents the spatiotemporal characteristics of distributed-parameter systems through spatial membership functions and fuzzy inference, and applied it to thermal-process modeling of battery modules [26]. Wu et al. introduced fuzzy membership-function and fuzzy-rule layers into PINNs, jointly constrained network training using physical-equation residuals and data errors, and applied the model to forward and inverse PDE problems involving fuzzy or imprecise data [27]. Ren and Wei combined fuzzy logic, neural networks, and physical constraints to predict the state of health of lithium-ion batteries [28]. These methods primarily introduce fuzzy logic at the levels of data representation, model fusion, or network architecture, whereas the physical processes or differential equations being solved remain deterministic.

Unlike the treatments of uncertainty at the data-representation or model-fusion level, PINNs have also begun to be used when equation parameters or parameter fields themselves involve interval or fuzzy uncertainty. Li et al. incorporated a fuzzy partial differential equation into a PINN as a physical constraint and combined it with long-term monitoring data to study pavement-performance degradation prediction [29]. Fuhg et al. proposed an interval physics-informed neural network (iPINN) for computing the bounds of interval PDE solutions with uncertain parameter fields, and assembled multiple iPINNs corresponding to different $\alpha$-cuts into a fuzzy physics-informed neural network (fPINN) to approximate the output of a fuzzy PDE [30]. These studies demonstrate that physics-informed learning can describe the effects of fuzzy or interval uncertainty on equation outputs. They also make the network representation of the unknown solution an issue requiring further consideration: when governing-equation constraints and uncertainty variables jointly enter the computational model, the selection of a trainable function-approximation architecture capable of describing the nonlinear relationships among space, time, and the uncertainty variables remains an open research problem.

Existing PINNs generally employ multilayer perceptrons (MLPs) to represent the unknown solutions of differential equations, and their training may be affected by gradient imbalance among loss terms and difficulties in learning high-frequency and multiscale features [31]-[33]. In addition to improving training strategies, modifying the functional representation of the unknown solution provides another research direction. Based on the Kolmogorov-Arnold representation theorem, Liu et al. proposed Kolmogorov-Arnold networks (KANs), in which the linear weights and fixed nodal activation functions of MLPs are replaced by learnable univariate functions defined on network connections [34]. Shukla et al. subsequently used KANs to construct PIKAN and DeepOKAN and compared KAN and MLP representations in forward and inverse differential-equation problems and operator-learning tasks [35]. Related studies have applied PIKAN to power-system dynamics prediction [36], deformation analysis of complex shell structures [37], flow in heterogeneous porous media [38], multi-material elasticity problems [39], and single-phase seepage equations with source and sink terms [40]. For fuzzy differential equations, Kazemi et al. combined PIKAN with granular fuzzy calculus based on relative-distance-measure fuzzy interval arithmetic to solve fuzzy fractional optimal-control problems, incorporating fractional-order dynamics, performance indices, observational data, and fuzzy-structure conditions such as inter-level nesting, crispness, and endpoint monotonicity into model training [41]. This study demonstrated one implementation of PIKAN for fuzzy fractional dynamical systems.

Under noisy intermediate-scale quantum (NISQ) computing conditions, variational quantum algorithms generally adopt a hybrid computational form that combines parameterized quantum circuits with classical optimizers [42]-[45]. Within this framework, a quantum neural network (QNN) constructs a mapping from inputs to outputs through data encoding, parameterized quantum-state transformations, and measurements, while a classical optimizer updates the circuit parameters. Previous studies have investigated the functional expressivity and model capacity of QNNs from the perspectives of data re-uploading, universal approximation, and effective dimension [46]-[48]. Havlíček et al. used quantum-enhanced feature spaces for supervised learning [49]; Zheng et al. constructed a quantum spatial graph convolutional network for graph-structured data [50]; Kyriienko et al. used differentiable parameterized quantum circuits to solve nonlinear differential equations [51]; and Rao et al. employed a QNN to predict carbon dioxide sequestration states in saline aquifers [52]. However, training parameterized quantum circuits may be affected by barren plateaus and quantum noise [53], [54]. Moreover, for differential-equation problems, the training objective of a general QNN does not automatically include the governing equations, initial conditions, or boundary conditions, and the corresponding physical information must therefore be incorporated into model training.

Following this idea, quantum physics-informed neural networks (QPINNs) incorporate governing-equation residuals, initial conditions, and boundary conditions into the training objective of a quantum model, thereby subjecting the parameterized quantum circuit to the corresponding physical conditions while it approximates the unknown solution. Markidis constructed a quantum PINN within a continuous-variable quantum-computing framework to solve a one-dimensional Poisson problem and examined the effects of the optimizer and quantum-network depth on the computational results [55]. Trahan et al. compared fully quantum, hybrid quantum-classical, and classical physics-informed networks for steady-state and transient differential equations [56]. Xiao et al. applied physics-informed quantum neural networks to forward and inverse problems involving partial differential equations [57]. Berger et al. introduced a trainable mapping into the quantum data-encoding stage and applied the resulting model to the Poisson, Burgers, and Navier-Stokes equations [58]. Panichi et al. further investigated the representation and solution of multivariable partial differential equations using QPINNs [59]. These studies connect quantum function approximation with differential-equation constraints, while also showing that model performance is influenced by quantum data encoding, circuit architecture, the number of available qubits, and the optimization process [55], [58].

To reduce the dependence of fully quantum models on fixed data encodings and deep quantum circuits, while exploiting the respective functional representations of classical networks and parameterized quantum circuits, some studies have constructed physics-informed models using both classical and quantum modules. Fernandez et al. embedded a quantum layer into a classical PINN and applied it to the wave equation [60]. Farea et al. proposed a quantum-classical physics-informed neural network (QCPINN) and compared different quantum-circuit configurations with classical PINNs in several PDE examples [61]. Leong et al. used a parallel hybrid architecture consisting of a quantum circuit and a classical network to solve high-speed-flow problems [62]. Song et al. investigated a hybrid quantum-classical solution of the incompressible Navier-Stokes equations on noisy quantum hardware [63]. Dehaghani et al. applied a hybrid quantum-classical physics-informed network to quantum optimal-control problems [64], whereas Rao et al. used QCPINN for porous-media flow equations [65]. Besides, Rao and Shen developed a first quantum-classical physics-informed Kolmogorov-Arnold network (QCPIKAN) [66]. Collectively, these studies have established a hybrid computational route in which classical modules perform part of the feature mapping, parameterized quantum circuits participate in the functional representation, and the entire model is jointly trained using a physics-informed loss.

The above studies have extended physics-informed learning models from different perspectives. PINNs based on $\alpha$-cuts provide trainable function representations for interval and fuzzy partial differential equations; PIKAN introduces KANs into forward and inverse differential-equation problems and has been further applied to fuzzy fractional systems; and QCPINN combines parameterized quantum circuits with classical physics-informed networks for various deterministic partial differential equations. For an FPDE represented by $\alpha$-cuts, the unknown comprises a family of interval-endpoint functions that varies with the membership level $\alpha$. In addition to satisfying the governing equation and initial-boundary conditions, its approximation involves endpoint ordering at the same $\alpha$ level and interval nesting across different $\alpha$ levels. For the triangular fuzzy quantities considered in this study, the endpoint-coincidence condition at $\alpha = 1$ must also be enforced. It is therefore worthwhile to investigate how a trainable model can jointly represent the effects of spatiotemporal variables and membership levels on the lower and upper endpoints while integrating KAN-based functional representations, parameterized quantum circuits, and fuzzy-structure constraints into a unified physics-informed computational framework.

To address this issue, this study constructs a QCPIKAN for solving fuzzy partial differential equations represented by $\alpha$-cuts. The model takes the spatiotemporal coordinates and membership level $\alpha$ as joint inputs and uses a ChebyKAN preprocessing module, a parameterized quantum circuit, and a ChebyKAN postprocessing module to form a hybrid function approximator that simultaneously outputs the lower and upper endpoints of the corresponding $\alpha$-cut solution. In addition to the governing-equation residuals and the associated initial-boundary conditions for the lower and upper endpoints, the loss function incorporates endpoint ordering, inter-level nesting enforced through endpoint monotonicity with respect to $\alpha$, and endpoint coincidence at $\alpha = 1$. A theoretical error analysis is further developed to compare QCPIKAN and PIKAN. Under the assumptions of well-posedness and residual stability, the endpoint-solution error is decomposed into approximation, optimization, finite-sampling, fuzzy-structure constraint, gradient-computation, and quantum-related error components. The resulting bound shows that QCPIKAN can achieve a smaller a priori error bound than PIKAN only when the representation gain provided by its hybrid feature space is sufficient to offset the differences in the remaining error components. Four numerical examples are considered: a fuzzy Poisson equation, a fuzzy heat-conduction equation, a fuzzy reaction-diffusion equation, and a fuzzy convection equation. Together, these examples cover elliptic, parabolic, and hyperbolic FPDEs and are used to evaluate endpoint-approximation accuracy, training-loss convergence, and the structural consistency of the resulting interval-solution families.

## 2 Methodology

### 2.1 $\alpha$-Cut Representation of Fuzzy Partial Differential Equations

This section presents the $\alpha$-cut representation of the FPDEs considered in this study and specifies the structural conditions that the corresponding interval-endpoint solutions must satisfy. Consider the spatial domain $\Omega \subset \mathbb{R}^d$ and a fuzzy-valued function $\tilde{u} = \tilde{u}(\boldsymbol{x}, t)$ defined on it. Its general governing equation and initial-boundary conditions can be expressed as

$$\begin{cases} \mathcal{N}\left[\tilde{u}; \tilde{\xi}\right] = \tilde{f}, \boldsymbol{x} \in \Omega, 0 < t \le T, \\ \mathcal{I}\left[\tilde{u}\right] = \tilde{u}_0, \boldsymbol{x} \in \Omega, t = 0, \\ \mathcal{B}\left[\tilde{u}\right] = \tilde{g}, \boldsymbol{x} \in \partial\Omega, 0 < t \le T, \end{cases} \tag{1}$$

where $\mathcal{N}$ is the differential operator associated with the governing equation; $\mathcal{I}$ and $\mathcal{B}$ denote the initial-condition and boundary-condition operators, respectively; $\tilde{\xi}$ denotes the fuzzy parameters in the equation; and $\tilde{f}$, $\tilde{u}_0$, and $\tilde{g}$ denote the potentially fuzzy source term, initial data, and boundary data, respectively. Deterministic parameters or data may be regarded as degenerate fuzzy quantities whose lower and upper endpoints coincide. For steady-state problems, the time variable and initial condition in Eq. (1) are omitted accordingly.

Let $\mu_{\tilde{\xi}}$ be the membership function of the fuzzy number $\tilde{\xi}$. For a prescribed membership level $\alpha \in (0,1]$, its $\alpha$-cut is defined as

$$\left[\tilde{\xi}\right]_\alpha = \left\{ z \in \mathbb{R} \mid \mu_{\tilde{\xi}}(z) \ge \alpha \right\}. \tag{2}$$

For a normal and convex fuzzy number, Eq. (2) is a closed interval and can be represented using lower- and upper-endpoint functions as follows [7]:

$$\left[\tilde{\xi}\right]_\alpha = \left[\underline{\xi}(\alpha), \overline{\xi}(\alpha)\right], \tag{3}$$

where $\underline{\xi}(\alpha)$ and $\overline{\xi}(\alpha)$ are the lower and upper endpoints of the $\alpha$-cut, respectively. The lower endpoint is nondecreasing with respect to $\alpha$, whereas the upper endpoint is nonincreasing, thereby ensuring inter-level nesting and contraction of the interval as the membership level increases [7]. All uncertain parameters in the numerical examples of this study are represented by triangular fuzzy numbers. Let $\tilde{\xi} = (\xi_l, \xi_m, \xi_r)$ and $\xi_l \le \xi_m \le \xi_r$. The corresponding $\alpha$-cut is then

$$\left[\tilde{\xi}\right]_{\alpha}=\left[\xi_l+\alpha\left(\xi_m-\xi_l\right),\xi_r-\alpha\left(\xi_r-\xi_m\right)\right]. \tag{4}$$

When $\alpha = 0$, Eq. (4) gives the support interval $[\xi_l, \xi_r]$ of the triangular fuzzy number. As $\alpha$ increases, this interval gradually contracts; when $\alpha = 1$, both endpoints are equal to $\xi_m$.

The fuzzy parameters and fuzzy solutions considered in this study are represented using $\alpha$-cuts because all uncertain parameters are normal, convex triangular fuzzy numbers whose uncertainty ranges and membership structures can be directly characterized by lower- and upper-endpoint functions. Moreover, the endpoint representation corresponds naturally to the governing equation, initial-boundary conditions, and dual-output architecture of the physics-informed model. The membership level $\alpha$ is further treated as a continuous input, enabling a single model to jointly approximate the lower- and upper-endpoint solutions at different $\alpha$ levels.

Accordingly, the fuzzy solution $\tilde{u}(\boldsymbol{x},t)$ at membership level $\alpha$ is represented as

$$\left[\tilde{u}(\boldsymbol{x},t)\right]_{\alpha}=\left[\underline{u}(\boldsymbol{x},t,\alpha),\overline{u}(\boldsymbol{x},t,\alpha)\right]. \tag{5}$$

Substituting the $\alpha$-cuts of the fuzzy parameters, source term, and initial-boundary data into Eq. (1) yields an interval problem at the prescribed $\alpha$ level. For the forms of parameter dependence considered in the examples of this study, the governing equations corresponding to the lower and upper endpoints can be written in the general form

$$\begin{cases}\mathcal{R}_L\left[\underline{u},\overline{u};\boldsymbol{x},t,\alpha\right]=0,\\ \mathcal{R}_U\left[\underline{u},\overline{u};\boldsymbol{x},t,\alpha\right]=0,\end{cases} \tag{6}$$

where $\mathcal{R}_L$ and $\mathcal{R}_U$ denote the governing-equation residuals corresponding to the lower and upper endpoints, respectively. The specific forms of the endpoint equations depend on the structure of the governing operator, the interval of the fuzzy parameters, and the monotonicity of the relevant terms with respect to the parameters and solution; they are therefore specified separately for each numerical example.

The interval solutions obtained at different $\alpha$ levels must also satisfy the structural conditions of a fuzzy solution. For any $0 \leq \alpha_1 \leq \alpha_2 \leq 1$, the endpoint functions should satisfy

$$\underline{u}(\boldsymbol{x},t,\alpha_1)\leq\underline{u}(\boldsymbol{x},t,\alpha_2)\leq\overline{u}(\boldsymbol{x},t,\alpha_2)\leq\overline{u}(\boldsymbol{x},t,\alpha_1). \tag{7}$$

Eq. (7) simultaneously imposes the endpoint-ordering and inter-level nesting. At the same $\alpha$ level, the lower endpoint should not exceed the upper endpoint; as $\alpha$ increases, the lower endpoint should be nondecreasing and the upper endpoint should be nonincreasing. If the endpoint functions are differentiable with respect to $\alpha$, the corresponding monotonicity conditions can be written as

$$\frac{\partial\underline{u}}{\partial\alpha}\geq 0,\ \frac{\partial\overline{u}}{\partial\alpha}\leq 0. \tag{8}$$

For the triangular fuzzy parameters with singleton cores considered in this study, all fuzzy inputs reduce to their deterministic core values at $\alpha = 1$. If the corresponding deterministic problem has a unique solution, the 1-cut of the fuzzy solution should also degenerate to a singleton, i.e.,

$$\underline{u}(\boldsymbol{x},t,1)=\overline{u}(\boldsymbol{x},t,1). \tag{9}$$

Consequently, solving the FPDEs considered in this study requires approximate solutions that satisfy not only the lower- and upper-endpoint governing equations and the corresponding initial-boundary conditions, but also the fuzzy-structure constraints of endpoint ordering, inter-level nesting, and endpoint coincidence at $\alpha = 1$. Based on this representation, the next section introduces QCPIKAN and incorporates these requirements into the corresponding physics-informed loss terms and fuzzy-structure constraints.

## 2.2 QCPIKAN for Fuzzy Differential Equations

Based on the $\alpha$-cut representation introduced in the preceding section, QCPIKAN is used to jointly approximate the lower- and upper-endpoint solutions at different membership levels. Let $\boldsymbol{r}$ denote the vector of independent variables of the partial differential equation. For a one-dimensional problem involving time, $\boldsymbol{r}=[t,x]^{\mathrm{T}}$; for a two-dimensional steady-state problem, $\boldsymbol{r}=[x,y]^{\mathrm{T}}$. After treating the membership level as an additional coordinate, the model input is uniformly expressed as

$$\boldsymbol{s}=\left[\boldsymbol{r}^{\mathrm{T}}\ \alpha\right]^{\mathrm{T}}. \tag{10}$$

QCPIKAN simultaneously outputs the lower and upper endpoints of the corresponding $\alpha$ -cut solution:

$$\hat{\boldsymbol{u}}_{\boldsymbol{\theta}}(\boldsymbol{s})=\left[\hat{u}_L(\boldsymbol{s})\quad \hat{u}_U(\boldsymbol{s})\right]^{\mathrm{T}}, \tag{11}$$

where $\hat{u}_L$ and $\hat{u}_U$ denote the network approximations of the lower and upper endpoints, respectively, and $\boldsymbol{\theta}$ denotes all trainable model parameters. The overall model mapping consists sequentially of a ChebyKAN preprocessing module, a parameterized quantum circuit, and a ChebyKAN postprocessing module:

$$\hat{\boldsymbol{u}}_{\boldsymbol{\theta}}=\mathcal{C}_{\mathrm{out}}\circ\mathcal{Q}_{\boldsymbol{\theta}_q}\circ\mathcal{C}_{\mathrm{in}}(\boldsymbol{s}), \tag{12}$$

where $\mathcal{C}_{\mathrm{in}}$, $\mathcal{Q}_{\boldsymbol{\theta}_q}$, and $\mathcal{C}_{\mathrm{out}}$ denote the ChebyKAN preprocessing mapping, parameterized quantum mapping, and ChebyKAN postprocessing mapping, respectively, and $\boldsymbol{\theta}_q$ denotes the quantum-circuit parameters. The explicit form of each mapping is given below.

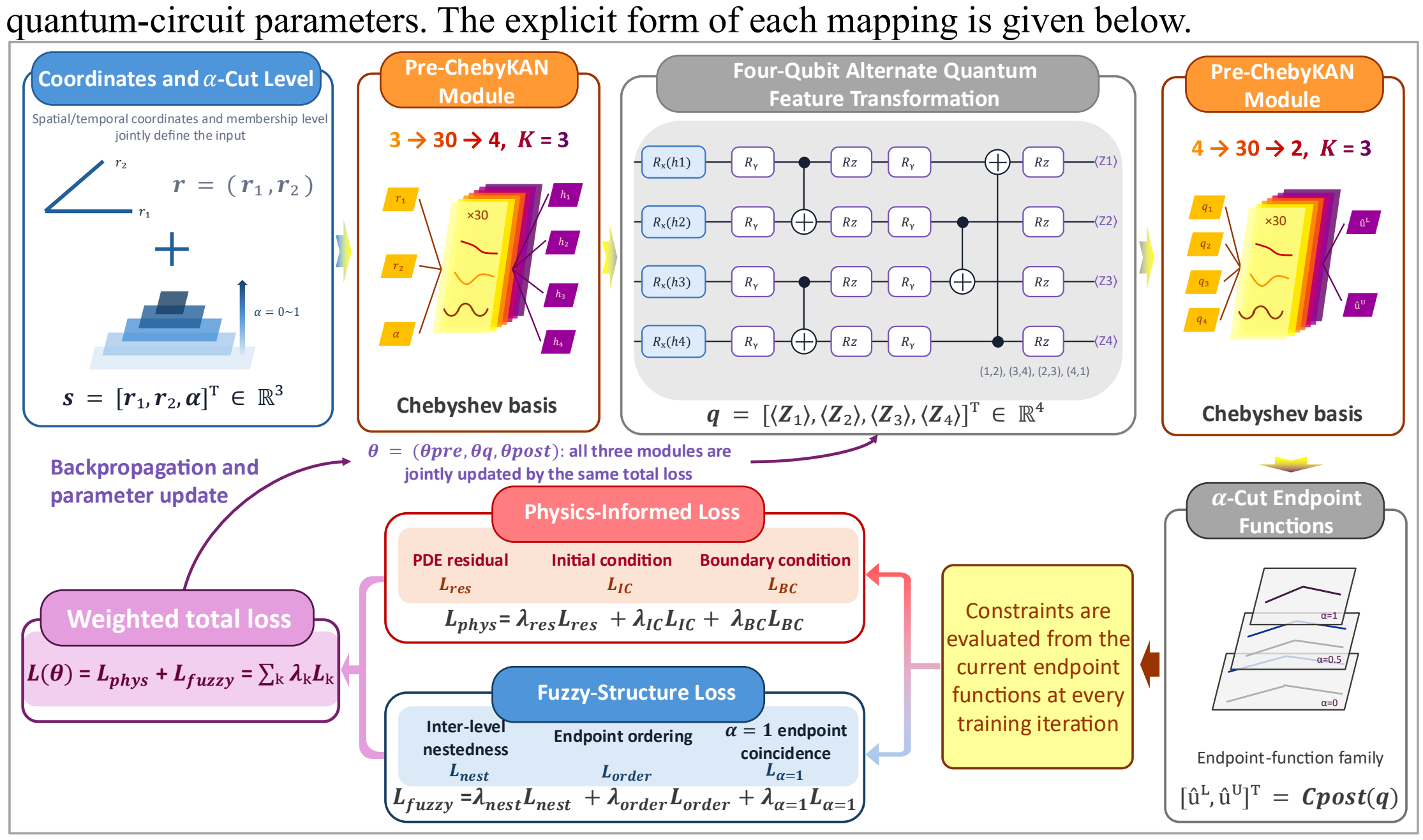

Fig. 1. Schematic of the QCPIKAN computational procedure for fuzzy differential equations.

### 2.2.1 ChebyKAN Function Representation

ChebyKAN layers are used to construct the classical mappings before and after the quantum circuit [67]. This architecture replaces the B-spline basis functions of the original KAN with first-kind Chebyshev polynomials and represents the univariate mapping between each input and output component using trainable expansion coefficients.

Let the input vector be $\boldsymbol{z}=\left[z_1\quad z_2\quad\cdots\quad z_{n_{\mathrm{in}}}\right]^{\mathrm{T}}$. Before evaluating the Chebyshev expansion, each input component is mapped to the interval $(-1,1)$ using the hyperbolic tangent function:

$$\bar{z}_i=\tanh(z_i),\quad i=1,2,\ldots,n_{\mathrm{in}}. \tag{13}$$

The first-kind Chebyshev polynomial of degree $k$ is defined as

$$T_k(\bar{z}_i)=\cos\left[k\arccos(\bar{z}_i)\right]. \tag{14}$$

Let $K$ be the maximum polynomial degree. The $j$ th output component of a ChebyKAN layer is then written as

$$y_j = \sum_{i=1}^{n_{\text{in}}}\sum_{k=0}^{K} c_{ijk} T_k\left(\bar{z}_i\right), \ j = 1, 2, \ldots, n_{\text{out}}, \tag{15}$$

where $n_{\text{in}}$ and $n_{\text{out}}$ denote the input and output dimensions of the layer, respectively, and $c_{ijk}$ denotes the trainable Chebyshev expansion coefficients. The ChebyKAN mapping formed by all output components is denoted by

$$\boldsymbol{y} = \mathcal{C}_{\boldsymbol{\Theta}}\left(\boldsymbol{z}\right), \tag{16}$$

where $\boldsymbol{\Theta}$ denotes the parameter set of the layer. The above ChebyKAN layers are used to construct the preprocessing and postprocessing mappings of the parameterized quantum circuit. Their specific connections are described in the next subsection.

#### 2.2.2 QCPIKAN Structure

The ChebyKAN preprocessing module maps the input vector $\boldsymbol{s}$ into a feature space whose dimension equals the number of qubits. Let the number of qubits be $n_q$. The preprocessing procedure is then expressed as

$$\boldsymbol{h} = \mathcal{C}_{\text{in}}\left(\boldsymbol{s}\right) = \mathcal{C}^{(2)}_{\boldsymbol{\Theta}_2}\left[\mathcal{C}^{(1)}_{\boldsymbol{\Theta}_1}\left(\boldsymbol{s}\right)\right], \boldsymbol{h} \in \mathbb{R}^{n_q}, \tag{17}$$

where $\boldsymbol{\Theta}_1$ and $\boldsymbol{\Theta}_2$ denote the parameters of the two preprocessing ChebyKAN layers, respectively.

The classical features $\boldsymbol{h}$ are embedded into a quantum state through angle encoding about the $X$ axis. Taking the all-zero state of $n_q$ qubits as the initial state, the encoded quantum state is

$$\left|\psi_{\text{enc}}\left(\boldsymbol{h}\right)\right\rangle = \left[\prod_{j=1}^{n_q} R_X^{(j)}\left(h_j\right)\right]\left|0\right\rangle^{\otimes n_q}. \tag{18}$$

The encoded quantum state is subsequently transformed by an alternate parameterized quantum circuit:

$$\left|\psi\left(\boldsymbol{h};\boldsymbol{\theta}_q\right)\right\rangle = U_{\text{alt}}\left(\boldsymbol{\theta}_q\right)\left|\psi_{\text{enc}}\left(\boldsymbol{h}\right)\right\rangle, \tag{19}$$

where $\boldsymbol{\theta}_q$ denotes the trainable parameters of the quantum circuit. The alternate circuit consists of parameterized two-qubit blocks acting on adjacent qubits. For control qubit $a$ and target qubit $b$, the block is expressed as

$$U_{a,b} = \left[R_Z^{(a)}\left(\theta_3\right) \otimes R_Z^{(b)}\left(\theta_4\right)\right]\text{CNOT}_{a\to b}\left[R_Y^{(a)}\left(\theta_1\right) \otimes R_Y^{(b)}\left(\theta_2\right)\right]. \tag{20}$$

This block sequentially applies two $R_Y$ rotations, one CNOT gate, and two $R_Z$ rotations. The CNOT gate introduces two-qubit coupling between adjacent qubits. Different blocks use staggered nearest-neighbor connections, allowing the qubits within the same variational layer to form a ring topology.

A Pauli-$Z$ measurement is performed separately on each qubit of the variational-circuit output state, yielding

$$q_j = \psi\left(\boldsymbol{h};\boldsymbol{\theta}_q\right)\right| Z_j \left|\psi\left(\boldsymbol{h};\boldsymbol{\theta}_q\right), \ j = 1, 2, \ldots, n_q. \tag{21}$$

All measured expectation values are assembled into the quantum feature vector

$$\boldsymbol{q} = \begin{bmatrix} q_1 & q_2 & \cdots & q_{n_q} \end{bmatrix}^{\text{T}} \in \left[-1, 1\right]^{n_q}. \tag{22}$$

The quantum features are mapped to the lower- and upper-endpoint outputs through the ChebyKAN postprocessing module:

$$\hat{\boldsymbol{u}}_{\boldsymbol{\theta}} = \mathcal{C}_{\text{out}}\left(\boldsymbol{q}\right) = \mathcal{C}^{(2)}_{\boldsymbol{\Phi}_2}\left[\mathcal{C}^{(1)}_{\boldsymbol{\Phi}_1}\left(\boldsymbol{q}\right)\right] = \left[\hat{u}_L \quad \hat{u}_U\right]^{\text{T}}, \tag{23}$$

where $\boldsymbol{\Phi}_1$ and $\boldsymbol{\Phi}_2$ denote the parameters of the two postprocessing ChebyKAN layers, respectively. All trainable model parameters are collectively denoted by

$$\boldsymbol{\theta}=\left\{\boldsymbol{\Theta}_1,\boldsymbol{\Theta}_2,\theta_q,\boldsymbol{\Phi}_1,\boldsymbol{\Phi}_2\right\}. \tag{24}$$

This study uses a three-dimensional input, 30-dimensional hidden features, and four qubits. The preprocessing module sequentially applies the ChebyKAN mappings $3\to30\to4$, whereas the postprocessing module sequentially applies the ChebyKAN mappings $4\to30\to2$. The parameterized quantum circuit contains one alternate variational layer. For four qubits, this layer acts on adjacent qubit pairs in the order $(1,2)$, $(3,4)$, $(2,3)$, and $(4,1)$, and contains four parameterized two-qubit blocks and 16 quantum-circuit parameters in total. The complete forward-propagation procedure can be expressed as

$$\mathbb{R}^3\to\mathbb{R}^{30}\to\mathbb{R}^4\to\mathcal{H}_2^{\otimes4}\to\mathcal{H}_2^{\otimes4}\to\mathbb{R}^4\to\mathbb{R}^{30}\to\mathbb{R}^2, \tag{25}$$

where the two occurrences of $\mathcal{H}_2^{\otimes4}$ denote the four-qubit state spaces after angle encoding and after transformation by the parameterized quantum circuit, respectively. The mapping from the quantum-state space to $\mathbb{R}^4$ is implemented through Pauli- $Z$ expectation-value measurements.

In the differentiable quantum-simulation environment used in this study, derivatives of the measurement expectation values with respect to the input features and quantum-circuit parameters are computed by backpropagation. Consequently, the classical ChebyKAN parameters and quantum-circuit parameters can be jointly updated under a unified physics-informed objective function.

### 2.2.3 Physics-Informed and Fuzzy-Structure Loss Functions

According to Eq. (23), for a prescribed space-time coordinate $\boldsymbol{r}$ and membership level $\alpha$, QCPIKAN simultaneously outputs the $\alpha$-cut solution, consisting of the lower endpoint $\hat{u}_L(\boldsymbol{r},\alpha;\boldsymbol{\theta})$ and upper endpoint $\hat{u}_U(\boldsymbol{r},\alpha;\boldsymbol{\theta})$, where $\boldsymbol{\theta}$ denotes all trainable model parameters. To ensure that the endpoint approximations simultaneously satisfy the governing equation, initial-boundary conditions, and fuzzy-structure requirements, the training loss comprises physical and fuzzy-structure constraint terms.

Let $\left\{\left(\boldsymbol{r}_r^i,\alpha_r^i\right)\right\}_{i=1}^{N_r}$ denote the residual collocation points, and let $\mathcal{R}_L^i$ and $\mathcal{R}_U^i$ denote the residuals obtained by substituting the lower- and upper-endpoint approximations into the corresponding $\alpha$-cut endpoint equations. The equation-residual loss is defined as

$$\mathcal{L}_{\text{res}}=\frac{1}{N_r}\sum_{i=1}^{N_r}\left[\left|\mathcal{R}_L^i\right|^2+\left|\mathcal{R}_U^i\right|^2\right], \tag{26}$$

where the derivatives of the endpoint functions with respect to the spatial and temporal variables are computed using automatic differentiation. The explicit endpoint residuals are determined jointly by the governing equation and the $\alpha$-cuts of the fuzzy parameters in the corresponding example.

For a time-dependent problem, let $\left\{\left(\boldsymbol{r}_{\text{IC}}^i,\alpha_{\text{IC}}^i\right)\right\}_{i=1}^{N_{\text{IC}}}$ denote the initial-condition sampling points and let $u_{0,L}^i$ and $u_{0,U}^i$ denote the corresponding initial endpoint values. The initial-condition loss is then

$$\mathcal{L}_{\text{IC}}=\frac{1}{N_{\text{IC}}}\sum_{i=1}^{N_{\text{IC}}}\left[\left|\hat{u}_L\left(\boldsymbol{r}_{\text{IC}}^i,\alpha_{\text{IC}}^i;\boldsymbol{\theta}\right)-u_{0,L}^i\right|^2+\left|\hat{u}_U\left(\boldsymbol{r}_{\text{IC}}^i,\alpha_{\text{IC}}^i;\boldsymbol{\theta}\right)-u_{0,U}^i\right|^2\right]. \tag{27}$$

Suppose that the problem contains $K$ boundary portions on which conditions must be imposed and that the $k$th boundary contains $N_{\text{BC}}^k$ sampling points. Let the target endpoint values be denoted by $g_{L,k}^i$ and $g_{U,k}^i$, respectively. The boundary-condition loss is defined as

$$\mathcal{L}_{\text{BC}}=\sum_{k=1}^{K}\frac{1}{N_{\text{BC}}^k}\sum_{i=1}^{N_{\text{BC}}^k}\left[\left|\hat{u}_L\left(\boldsymbol{r}_{\text{BC},k}^i,\alpha_{\text{BC},k}^i;\boldsymbol{\theta}\right)-g_{L,k}^i\right|^2+\left|\hat{u}_U\left(\boldsymbol{r}_{\text{BC},k}^i,\alpha_{\text{BC},k}^i;\boldsymbol{\theta}\right)-g_{U,k}^i\right|^2\right]. \tag{28}$$

In addition to the above physical constraints, the two endpoint functions in Eq. (23) must jointly describe a consistent fuzzy-valued solution. According to the level-set structure introduced in Section 2.1, as $\alpha$ increases, the lower endpoint should be nondecreasing and the upper endpoint should be nonincreasing. Let $[z]_+ = \max(0, z)$ denote the positive-part function and let $\left\{\left(\boldsymbol{r}_n^i, \alpha_n^i\right)\right\}_{i=1}^{N_{\text{nest}}}$ denote the nesting collocation points. The inter-level nesting loss is defined as

$$\mathcal{L}_{\text{nest}} = \frac{1}{N_{\text{nest}}} \sum_{i=1}^{N_{\text{nest}}} \left\{ \left[ -\frac{\partial \hat{u}_L}{\partial \alpha}\left(\boldsymbol{r}_n^i, \alpha_n^i; \boldsymbol{\theta}\right) \right]_+ + \left[ \frac{\partial \hat{u}_U}{\partial \alpha}\left(\boldsymbol{r}_n^i, \alpha_n^i; \boldsymbol{\theta}\right) \right]_+ \right\}, \tag{29}$$

where only endpoint derivatives opposite to the expected monotonic direction are penalized, and the derivatives with respect to $\alpha$ are likewise obtained by automatic differentiation.

The inter-level nesting loss constrains the nesting relation among different $\alpha$ levels. At each fixed $\alpha$ level, the lower endpoint must additionally not exceed the upper endpoint. Let $\left\{\left(\boldsymbol{r}_o^i, \alpha_o^i\right)\right\}_{i=1}^{N_{\text{order}}}$ denote the endpoint-ordering collocation points. The corresponding endpoint-ordering loss is defined as

$$\mathcal{L}_{\text{order}} = \frac{1}{N_{\text{order}}} \sum_{i=1}^{N_{\text{order}}} \left[ \hat{u}_L\left(\boldsymbol{r}_o^i, \alpha_o^i; \boldsymbol{\theta}\right) - \hat{u}_U\left(\boldsymbol{r}_o^i, \alpha_o^i; \boldsymbol{\theta}\right) \right]_+ . \tag{30}$$

For the triangular fuzzy parameters with singleton cores considered in this study, the parameter interval reduces to a deterministic value at $\alpha = 1$, and the lower and upper endpoints of the solution should coincide. Accordingly, the endpoint-coincidence loss at $\alpha = 1$ is defined as follows:

$$\mathcal{L}_{\alpha=1} = \frac{1}{N_{\alpha=1}} \sum_{i=1}^{N_{\alpha=1}} \left| \hat{u}_L\left(\boldsymbol{r}_c^i, 1; \boldsymbol{\theta}\right) - \hat{u}_U\left(\boldsymbol{r}_c^i, 1; \boldsymbol{\theta}\right) \right|^2 , \tag{31}$$

this term enforces the endpoint-coincidence condition at $\alpha = 1$, under which the corresponding solution $\alpha$ -cut reduces to a singleton.

Combining the above terms, the total loss function is written as

$$L(\boldsymbol{\theta}) = \lambda_{\text{res}}\mathcal{L}_{\text{res}} + \lambda_{\text{IC}}\mathcal{L}_{\text{IC}} + \lambda_{\text{BC}}\mathcal{L}_{\text{BC}} + \lambda_{\text{nest}}\mathcal{L}_{\text{nest}} + \lambda_{\text{order}}\mathcal{L}_{\text{order}} + \lambda_{\alpha=1}\mathcal{L}_{\alpha=1} , \tag{32}$$

where all weighting coefficients are nonnegative constants used to balance the relative contributions of the different loss components during training. For a steady-state problem without an initial condition, set $\lambda_{\text{IC}} = 0$ . The QCPIKAN parameters are obtained by solving the following optimization problem:

$$\boldsymbol{\theta}^* = \arg\min_{\boldsymbol{\theta}} \mathcal{L}(\boldsymbol{\theta}) . \tag{33}$$

Eqs. (29)-(31) penalize violations of the fuzzy-structure constraints through soft enforcement. Therefore, endpoint ordering and inter-level nesting cannot be assumed to hold strictly throughout the continuous domain solely on the basis of these loss definitions. In the numerical experiments, the corresponding fuzzy-structure violations are further evaluated at independently sampled test points. The sampling sizes and loss weights used in each example are specified separately in the numerical-experiment section.

### 2.3 Theoretical Error Advantage of QCPIKAN over PIKAN

This section analyzes the conditions under which QCPIKAN may achieve higher accuracy than PIKAN in terms of function-approximation error, physical-residual error, and fuzzy-structure constraint error. It must be emphasized that a parameterized quantum circuit cannot unconditionally guarantee higher accuracy for an arbitrary fuzzy differential equation. The following results are therefore established under the assumptions that the equation is well posed and stable, both models use the same physics-informed loss, quantum-measurement errors are negligible, and the QCPIKAN feature space can effectively represent nonseparable couplings among the input variables. This formulation is consistent with the understanding from studies of universal approximation and effective dimension in quantum neural networks and quantum physics-informed networks that final accuracy is jointly determined by model capacity and training error [47], [48], [55], [58].

Let $z=(x,t,\alpha)$ denote the joint input, let $U(z)=\left(u^{-}(z),u^{+}(z)\right)$ denote the exact lower- and upper-endpoint solution of the fuzzy differential equation, and let $\mathcal{D}$ denote the corresponding spatiotemporal-membership domain. For any model $M\in\{P,Q\}$, let $\mathcal{V}_P$ and $\mathcal{V}_Q$ denote the function-representation spaces of PIKAN and QCPIKAN, respectively, and define the best-approximation error as

$$E_M^{\text{app}}=\inf_{V\in\mathcal{V}_M}\|U-V\|_H,\ M\in\{P,Q\},\tag{34}$$

here, $\|\cdot\|_H$ may be defined over $\mathcal{D}$ as a weighted $L^2$ norm of the endpoint functions or may be extended to a Sobolev norm containing derivatives with respect to space, time, or membership level. This definition distinguishes errors caused by the representational capacity of the network from those arising during training. For the $\alpha$-cut representation used in this study, the $H$ norm should act on both the lower and upper endpoints, thereby reflecting the approximation quality of the entire fuzzy-solution family rather than only the pointwise error at a fixed $\alpha$ level.

Let $\mathcal{A}$ denote the joint operator comprising the lower- and upper-endpoint governing equations, initial-boundary conditions, and the fuzzy-structure constraints of endpoint ordering, inter-level nesting, and endpoint coincidence at $\alpha=1$. If the problem satisfies a residual-to-solution error stability estimate, then there exists a network-parameter-independent constant $C_{\text{stab}}>0$ such that any approximate solution $V$ satisfies

$$\|U-V\|_H\le C_{\text{stab}}\|\mathcal{A}(V)-\mathcal{A}(U)\|_Y.\tag{35}$$

Eq. (35) shows that the physics-informed loss is not merely an optimization objective but also serves as a stability surrogate controlling the solution error. For the weighted loss in Eq. (32), if the weights of all loss terms are positive and the sampling error is controlled, the upper bound on the total error of model $M$ can be written as

$$\mathcal{E}_M\le C_{\text{stab}}\left(E_M^{\text{app}}+\varepsilon_M^{\text{opt}}+\varepsilon_M^{\text{sam}}+\varepsilon_M^{\text{str}}+\varepsilon_M^{\text{noise}}\right),\tag{36}$$

where $\mathcal{E}_M$ denotes the endpoint-solution error, $\varepsilon_M^{\text{opt}}$ the error caused by incomplete optimization convergence, $\varepsilon_M^{\text{sam}}$ the error arising from finite collocation and numerical integration, $\varepsilon_M^{\text{str}}$ the residual error associated with fuzzy-structure constraints such as endpoint ordering, inter-level nesting, and endpoint coincidence at $\alpha=1$, and $\varepsilon_M^{\text{noise}}$ the error introduced by quantum measurement or hardware noise. Under the setting adopted in this study, in which an ideal quantum simulator and analytic expectation values are used, $\varepsilon_Q^{\text{noise}}=0$; this term cannot be directly neglected on a real NISQ device.

The potential advantage of QCPIKAN arises from its hybrid feature space. If PIKAN is viewed as a model using only classical ChebyKAN mappings, whereas QCPIKAN additionally introduces a parameterized quantum feature mapping for the same inputs, the quantum measurement features can be written as

$$\phi_Q(z)=\left(0\middle|\otimes n_q E^{\dagger}(z)V_{\boldsymbol{\theta}}^{\dagger}P_jV_{\boldsymbol{\theta}}E(z)\middle|0^{\otimes n_q}\right)_{j=1}^{n_q},\tag{37}$$

where $E(z)$ denotes angle encoding, $V_{\boldsymbol{\theta}}$ denotes a parameterized quantum circuit containing entangling gates, and $P_j$ denotes the measurement operator. Compared with a classical representation that performs only variable-wise univariate mappings, an entangling circuit can introduce mixed terms among variables into the measured features. Consequently, when the fuzzy endpoint solution simultaneously contains space-time coupling, space-time-$\alpha$ coupling, or parameter-solution coupling, QCPIKAN can provide additional interaction features under the same or a similar classical parameter budget. Suppose that the exact solution is decomposed as $U=U_{\text{sep}}+U_{\text{int}}$, where $U_{\text{int}}$ denotes the aforementioned nonseparable component. If, for a prescribed model size,

$$\inf_{V\in\mathcal{V}_Q}\|U_{\text{int}}-V\|_H=\varepsilon_Q^{\text{int}}<\varepsilon_P^{\text{int}}=\inf_{V\in\mathcal{V}_P}\|U_{\text{int}}-V\|_H,\tag{38}$$

then QCPIKAN has a strict best-approximation advantage in representing the nonseparable structure of the fuzzy solution. This advantage is generally more pronounced at low to intermediate membership levels, in propagation regions with large gradients, and when fuzzy parameters have a substantial effect on the decay rate or wavefront position of the solution.

To give the above comparison a rigorous mathematical meaning, the model error is decomposed below into representation and computational errors, and a sufficient condition is provided under which QCPIKAN has a smaller theoretical error bound. Here, "higher accuracy" means a smaller a priori error bound under the same error norm. This bound advantage will manifest directly as an actual accuracy advantage in numerical experiments only when the error bound is sufficiently tight and training is adequate.

Let $\hat{U}_M$ be the approximation obtained through training. The theoretical analysis employs the following assumptions. A1: the endpoint problem is well posed and satisfies the residual-stability condition in Eq. (35). A2: the two models use the same physical operator, fuzzy-structure constraints on the $\alpha$-cut endpoints, loss weights, sampling rules, and error norm. A3: the hybrid QCPIKAN feature space can at least reproduce the classical PIKAN representation or has a strict best-approximation advantage at a fixed model size. A4: the sampling error caused by finite collocation, the error caused by incomplete optimization convergence, and the error associated with soft fuzzy structural constraints can all be bounded by finite terms. In particular, the numerical experiments use analytic expectation values in an ideal quantum-simulation environment and calculate gradients with respect to the network inputs and parameters through automatic differentiation or analytic backpropagation. They therefore contain neither stochastic measurement noise nor stochastic gradient noise, i.e., $\varepsilon_P^{\text{grad}} = \varepsilon_Q^{\text{grad}} = 0$. Moreover, ideal quantum simulation introduces no hardware noise, i.e., $\varepsilon_Q^{\text{hw}} = 0$. These error terms must be retained for real quantum hardware.

Under Assumption A3, if the PIKAN representation space is embedded in the QCPIKAN representation space, then

$$\mathcal{V}_P \subseteq \mathcal{V}_Q, E_Q^{\text{app}} \le E_P^{\text{app}}. \tag{39}$$

Eq. (39) gives a non-inferiority result arising from a non-strict inclusion relation. A strict advantage additionally requires that the target solution contains a nonseparable interaction component that is difficult for PIKAN to represent at the prescribed network size but can be effectively approximated by the quantum feature mapping of QCPIKAN. Let

$$U = U_{\text{sep}} + U_{\text{int}} + U_{\text{res}}, \left\| U_{\text{int}} - \Pi_Q U_{\text{int}} \right\|_H < \left\| U_{\text{int}} - \Pi_P U_{\text{int}} \right\|_H, \tag{40}$$

where $U_{\text{sep}}$ denotes the separable component that is readily represented by the classical branch, $U_{\text{int}}$ denotes the mixed component among space, time, and membership level, $U_{\text{res}}$ denotes the remaining approximation error, and $\Pi_P$ and $\Pi_Q$ denote the best projections onto the two feature spaces, respectively. Eq. (40) states that the quantum entanglement features approximate the nonseparable dependence among the spatial, temporal, and membership-level variables more accurately than the PIKAN representation. Consequently, QCPIKAN has a positive representation gain $\Delta_{\text{app}} = E_P^{\text{app}} - E_Q^{\text{app}}$ in the error decomposition. This condition is not automatically guaranteed by the presence of a quantum circuit; rather, it is jointly determined by the circuit architecture, input encoding, model size, and interaction complexity of the target solution.

The complete comparison of the error bounds is given below. Define

$$\varepsilon_M = \varepsilon_M^{\text{opt}} + \varepsilon_M^{\text{sam}} + \varepsilon_M^{\text{str}} + \varepsilon_M^{\text{grad}} + \varepsilon_M^{\text{hw}}, \bar{\mathcal{E}}_M = C_{\text{stab}} \left( E_M^{\text{app}} + \varepsilon_M \right), \tag{41}$$

where $\varepsilon_M^{\text{opt}}$ denotes the optimization error, $\varepsilon_M^{\text{sam}}$ the finite-collocation and numerical-integration error, $\varepsilon_M^{\text{str}}$ the residual error of fuzzy structural constraints such as endpoint ordering, inter-level nesting, and endpoint coincidence at $\alpha = 1$, $\varepsilon_M^{\text{grad}}$ the gradient-computation error, $\varepsilon_M^{\text{hw}}$ the quantum-hardware noise error, and $\bar{\mathcal{E}}_M$ the theoretical error bound. By the triangle inequality, the error between the exact and trained solutions can be decomposed into the best-representation error and the training-process error. Combining the stability estimate in Eq. (35) with the control of the residuals by the loss terms in Eq. (32) yields Eq. (42). Under the ideal conditions considered here, $\varepsilon_P^{\text{grad}} = \varepsilon_Q^{\text{grad}} = \varepsilon_Q^{\text{hw}} = 0$. Because PIKAN involves no quantum-hardware noise, $\varepsilon_P^{\text{hw}} = 0$. It should be noted that the finite-collocation error $\varepsilon_M^{\text{sam}}$ and optimization error $\varepsilon_M^{\text{opt}}$ do not automatically vanish merely because the gradients are evaluated analytically.

$$\mathcal{E}_M = \left\| U - \hat{U}_M \right\|_H \le \bar{\mathcal{E}}_M \text{ , } M \in \{P, Q\} . \tag{42}$$

Theorem 1. Under Assumptions A1-A4, if the representation gain of QCPIKAN exceeds its additional computational error relative to PIKAN, i.e.,

$$\Delta_{\text{app}} > \left(\varepsilon_Q^{\text{opt}} - \varepsilon_P^{\text{opt}}\right) + \left(\varepsilon_Q^{\text{sam}} - \varepsilon_P^{\text{sam}}\right) + \left(\varepsilon_Q^{\text{str}} - \varepsilon_P^{\text{str}}\right) + \left(\varepsilon_Q^{\text{grad}} - \varepsilon_P^{\text{grad}}\right) + \left(\varepsilon_Q^{\text{hw}} - \varepsilon_P^{\text{hw}}\right), \tag{43}$$

for the ideal numerical experiments considered in this study, both the gradient error and hardware-noise error in Eq. (43) are zero. The sufficient condition of the theorem therefore reduces to

$$\Delta_{\text{app}} > \left(\varepsilon_Q^{\text{opt}} - \varepsilon_P^{\text{opt}}\right) + \left(\varepsilon_Q^{\text{sam}} - \varepsilon_P^{\text{sam}}\right) + \left(\varepsilon_Q^{\text{str}} - \varepsilon_P^{\text{str}}\right), \tag{44}$$

then the theoretical endpoint-solution error bound of QCPIKAN is strictly smaller than that of PIKAN, i.e.,

$$\bar{\mathcal{E}}_Q < \bar{\mathcal{E}}_P . \tag{45}$$

Proof: First, from Eqs. (36), (41), and (42), the two models satisfy, respectively,

$$\mathcal{E}_Q \le \bar{\mathcal{E}}_Q = C_{\text{stab}}\left(E_Q^{\text{app}} + \varepsilon_Q\right), \mathcal{E}_P \le \bar{\mathcal{E}}_P = C_{\text{stab}}\left(E_P^{\text{app}} + \varepsilon_P\right). \tag{46}$$

Second, Eq. (43) is equivalent to

$$\left(E_Q^{\text{app}} + \varepsilon_Q\right) - \left(E_P^{\text{app}} + \varepsilon_P\right) < 0 . \tag{47}$$

Because the stability constant $C_{\text{stab}} > 0$, multiplying both sides of Eq. (47) by $C_{\text{stab}}$ does not change the direction of the inequality. Therefore,

$$C_{\text{stab}}\left(E_Q^{\text{app}} + \varepsilon_Q\right) < C_{\text{stab}}\left(E_P^{\text{app}} + \varepsilon_P\right). \tag{48}$$

Thus, $\bar{\mathcal{E}}_Q < \bar{\mathcal{E}}_P$ and Eq. (45) holds, completing the proof. For the ideal-gradient experiments considered here, $\varepsilon_P^{\text{grad}} = \varepsilon_Q^{\text{grad}} = \varepsilon_P^{\text{hw}} = \varepsilon_Q^{\text{hw}} = 0$, Therefore, Eqs. (43) and (44) are fully equivalent. The key point of the proof is that QCPIKAN does not automatically gain an advantage merely from the presence of a quantum module. Instead, it has a strictly smaller theoretical error bound only when the reduction in representation error, $\Delta_{\text{app}}$, is sufficient to offset the differences in optimization error, finite-sampling error, and fuzzy-structure constraint error.

For the fuzzy Poisson, fuzzy heat-conduction, fuzzy reaction-diffusion, and fuzzy convection equations considered in this study, $U_{\text{int}}$ primarily manifests as coupling between the spatiotemporal variables and $\alpha$. For example, a fuzzy diffusion coefficient changes the decay rate of the heat-conduction solution, whereas a fuzzy convection velocity changes the wavefront position. The endpoint functions are therefore generally not simple sums of a spatial function, a temporal function, and a membership-level function. A quantum feature mapping containing entangling gates can provide additional interaction features for these mixed variations, which explains why QCPIKAN may achieve lower errors at low to intermediate $\alpha$ levels, in high-gradient regions, and near wavefronts.

It must be emphasized that Theorem 1 rigorously guarantees a comparison of error bounds rather than $\mathcal{E}_Q < \mathcal{E}_P$ for all problems, random initializations, and quantum-hardware conditions. For the experiments in this study, which use analytic gradients and no quantum-measurement noise, the principal terms to be compared are the representation, finite-collocation, optimization, and soft-structure constraint errors. If finite-shot quantum measurements, numerically approximated gradients, or real quantum hardware are used, $\varepsilon_M^{\text{grad}}$ and $\varepsilon_M^{\text{hw}}$ may increase, and the simplified condition in Eq. (44) should revert to the more general condition in Eq. (43); the theoretical advantage may then be offset. The numerical results presented here should therefore be interpreted as empirical verification of Eq. (44) under ideal conditions.

## 3 Numerical Examples

To evaluate the ability of QCPIKAN to solve different types of fuzzy PDEs, numerical experiments are conducted for the fuzzy Poisson equation, fuzzy heat-conduction equation, fuzzy reaction-diffusion equation, and fuzzy convection equation, covering elliptic, parabolic, and hyperbolic problems. Except for the loss weights specified separately for each equation, QCPIKAN and PIKAN use the same training data, optimization parameters, and fuzzy-structure constraints. The training-iteration parameter is set to 20000 for all models, with a batch size of 256 and an initial learning rate of $5 \times 10^{-5}$. Within each example, QCPIKAN and PIKAN use the same sampling rules and sample sizes. After training, 50000 randomly generated test points are used to evaluate inter-level nesting violations in the predicted solution intervals. Predictions are generated at five membership levels, $\alpha = 0, 0.25, 0.50, 0.75$, and $1.00$. QCPIKAN uses four qubits and one alternate parameterized quantum-circuit layer, and the quantum measurement outputs are evaluated as analytic expectation values in an ideal quantum-simulation environment.

### 3.1 Example 1: Elliptic Fuzzy Poisson Equation

To test the ability of QCPIKAN to solve a two-dimensional elliptic FPDE, a Poisson boundary-value problem driven by a fuzzy source term is considered as the first example. Let $\Omega = (0,1)^2$. Over the domain $\Omega$, seek the fuzzy-valued function $\tilde{u}(x,y)$ such that

$$\begin{cases} -\Delta \tilde{u}(x,y) = \tilde{f}, (x,y) \in \Omega, \\ \tilde{u}(x,y) = 0, (x,y) \in \partial\Omega, \end{cases} \tag{49}$$

where $\Delta = \partial^2 / \partial x^2 + \partial^2 / \partial y^2$ is the spatial Laplace operator and the source term is the triangular fuzzy number $\tilde{f} = (0.5, 1.0, 1.5)$. The problem has homogeneous Dirichlet boundaries; its solution vanishes on the boundary and is positive in the interior. It can therefore be used to evaluate the joint representation of two-dimensional spatial morphology, fuzzy-interval propagation, and boundary constraints.

At membership level $\alpha \in [0,1]$, the $\alpha$-cut of the fuzzy source term is

$$\left[\tilde{f}\right]_\alpha = \left[f_\alpha^L, f_\alpha^U\right] = \left[0.5 + 0.5\alpha, 1.5 - 0.5\alpha\right]. \tag{50}$$

Accordingly, Eq. (49) can be transformed into the following lower- and upper-endpoint boundary-value problems:

$$\begin{cases} -\Delta u_\alpha^L(x,y) = f_\alpha^L, (x,y) \in \Omega, \\ -\Delta u_\alpha^U(x,y) = f_\alpha^U, (x,y) \in \Omega, \\ u_\alpha^L(x,y) = u_\alpha^U(x,y) = 0, (x,y) \in \partial\Omega. \end{cases} \tag{51}$$

In this example, both QCPIKAN and PIKAN take $(x, y, \alpha)$ as input and simultaneously approximate $u_\alpha^L$ and $u_\alpha^U$ in Eq. (36).

To quantitatively evaluate the model predictions, the Fourier sine-series solution of Eq. (51) is used as the reference solution. For endpoint index $\ell \in \{L, U\}$, its truncated form is

$$u_\alpha^\ell(x,y) = \sum_{\substack{1 \le m,n \le 49 \\ m,n\ \text{odd}}} \frac{16 f_\alpha^\ell}{\pi^4 mn\left(m^2 + n^2\right)} \sin(m\pi x)\sin(n\pi y), \ell \in \{L, U\}. \tag{52}$$

Eq. (52) is obtained from the eigenfunction expansion of the homogeneous Dirichlet Poisson problem on the unit square. The reference solution is evaluated on a $201 \times 201$ uniform spatial grid and interpolated onto the $50 \times 50$ model-prediction grid for error evaluation.

This example is a steady-state elliptic problem without a time variable and therefore requires no initial-condition loss, i.e., $\lambda_{\text{IC}} = 0$. Within the unified loss framework in Eq. (32), $\lambda_{\text{res}} = \lambda_{\text{BC}} = \lambda_{\alpha=1} = 1$, $\lambda_{\text{nest}} = 0.1$, and $\lambda_{\text{order}} = 0.2$ are used, yielding the total loss $L(\boldsymbol{\theta}) = \mathcal{L}_{\text{res}} + \mathcal{L}_{\text{BC}} + 0.1\mathcal{L}_{\text{nest}} + 0.2\mathcal{L}_{\text{order}} + \mathcal{L}_{\alpha=1}$. QCPIKAN and PIKAN use identical training points, loss weights, and optimization parameters to ensure comparison under consistent training conditions.

Figs. 2-4 compare the training behavior, endpoint fields, and fuzzy-structure properties of QCPIKAN and PIKAN. Both loss curves decrease throughout training; QCPIKAN falls below PIKAN after approximately 8000 epochs and finishes with about half the PIKAN loss. Both models reproduce the homogeneous boundary values and the central peak, with closely matching solution envelopes. The interval width is largest in the interior and decreases toward the boundary, although small negative widths remain locally. Inter-level nesting violations are concentrated near the boundary and corners; PIKAN has denser lower-endpoint violations, whereas the upper-endpoint violation patterns are similar for the two models.

Relative to the reference solution, Fig. 5 shows that the largest errors of both models are of the same order, but the PIKAN errors are more spatially widespread, especially for the upper endpoint. Fig. 6(a) shows that PIKAN is slightly more accurate at $\alpha = 0$; at $\alpha = 0.25$, 0.50, 0.75, and 1.00, its mean relative $L_2$ errors are approximately 1.4, 1.6, 1.9, and 2.0 times those of QCPIKAN, respectively. Fig. 6(b) confirms that both models capture interval contraction with increasing $\alpha$, while the QCPIKAN lower endpoint is generally closer to the reference curve and the PIKAN upper endpoint is overestimated at some levels. In Fig. 7, increasing the qubit count from two to six reduces the mean absolute error for a one-layer circuit; with four qubits, two layers give the lowest error, whereas a third layer degrades the result. Thus, accuracy improves with the tested qubit counts but varies nonmonotonically with circuit depth. This observation is consistent with the discoveries of Rao et al. [65]. As circuit depth keeps increasing, accuracy may stop rising and even deteriorate, which can be explained by growing optimization difficulty and the barren plateau effect crippling gradient-based training of variational quantum circuits.

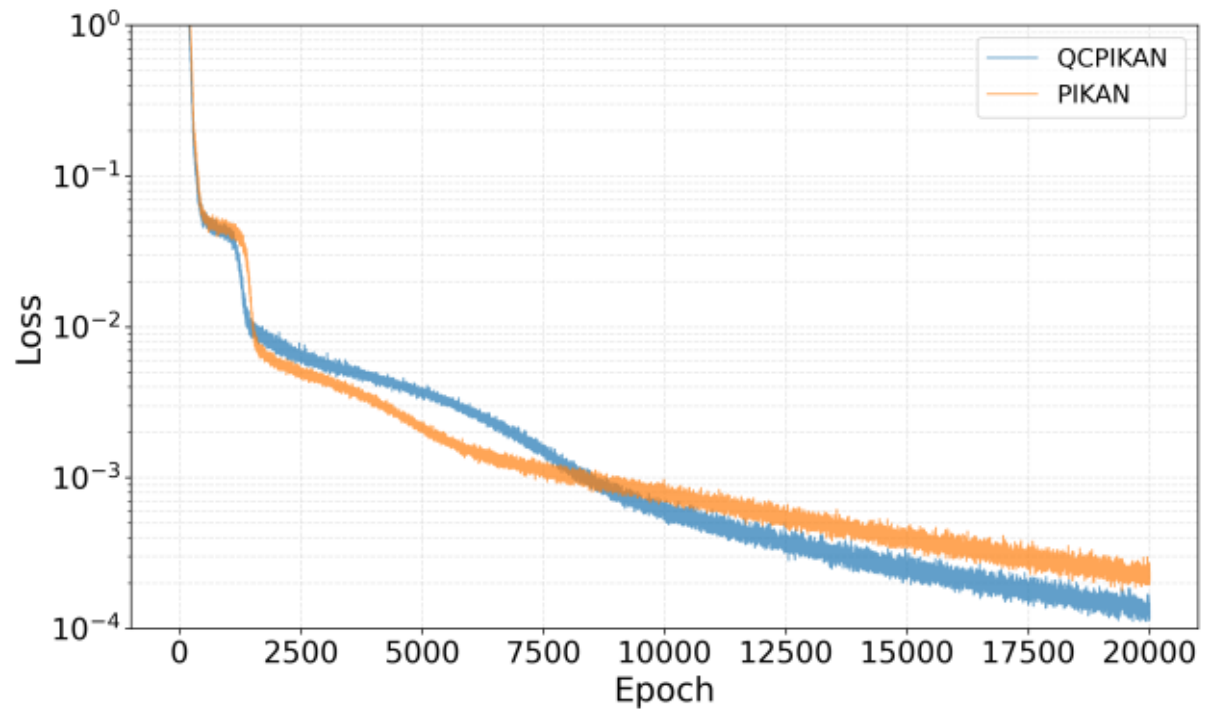

Fig. 2. Comparison of training losses in Example 1.

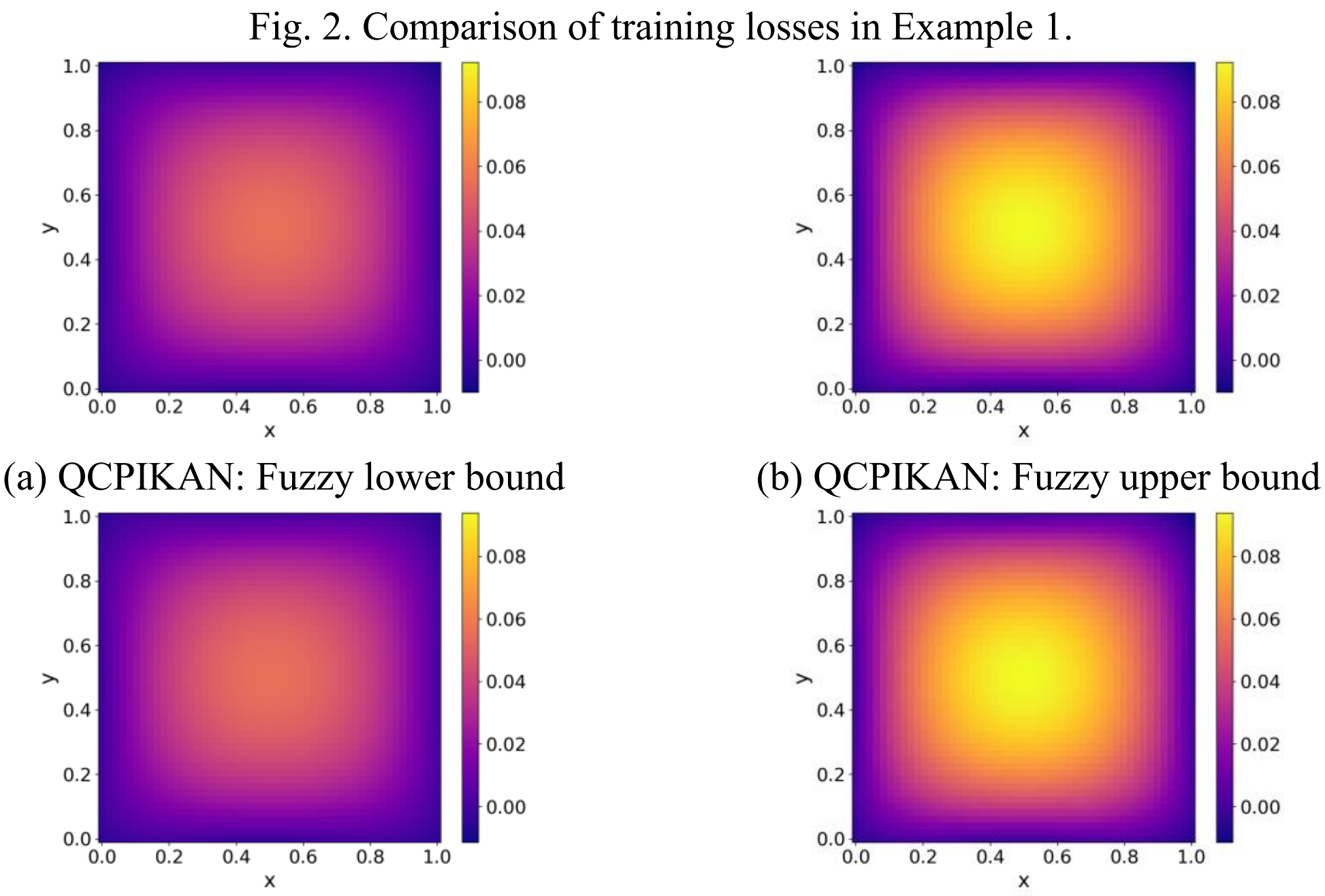

(a) QCPIKAN: Fuzzy lower bound

(b) QCPIKAN: Fuzzy upper bound

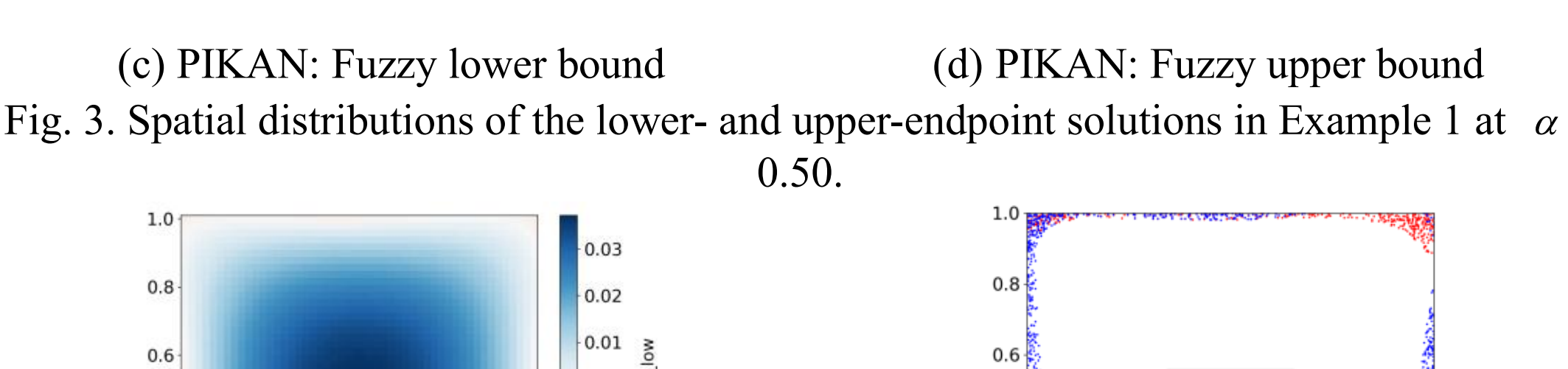

(c) PIKAN: Fuzzy lower bound (d) PIKAN: Fuzzy upper bound

Fig. 3. Spatial distributions of the lower- and upper-endpoint solutions in Example 1 at $\alpha$ = 0.50.

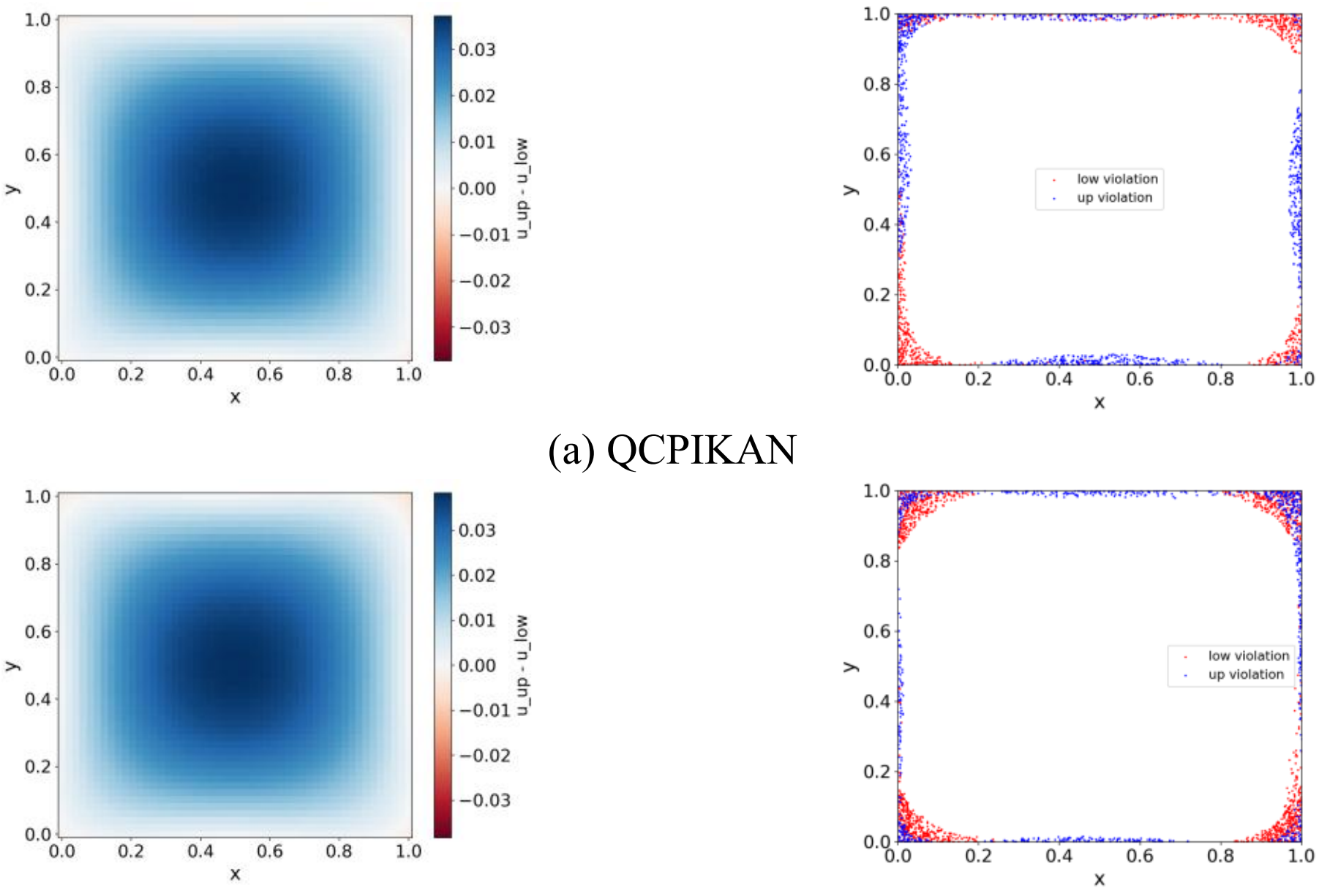

(a) QCPIKAN

(b) PIKAN

Fig. 4. Spatial interval width and inter-level nesting violations along the $\alpha$ direction at $\alpha$ = 0.5 in Example 1.

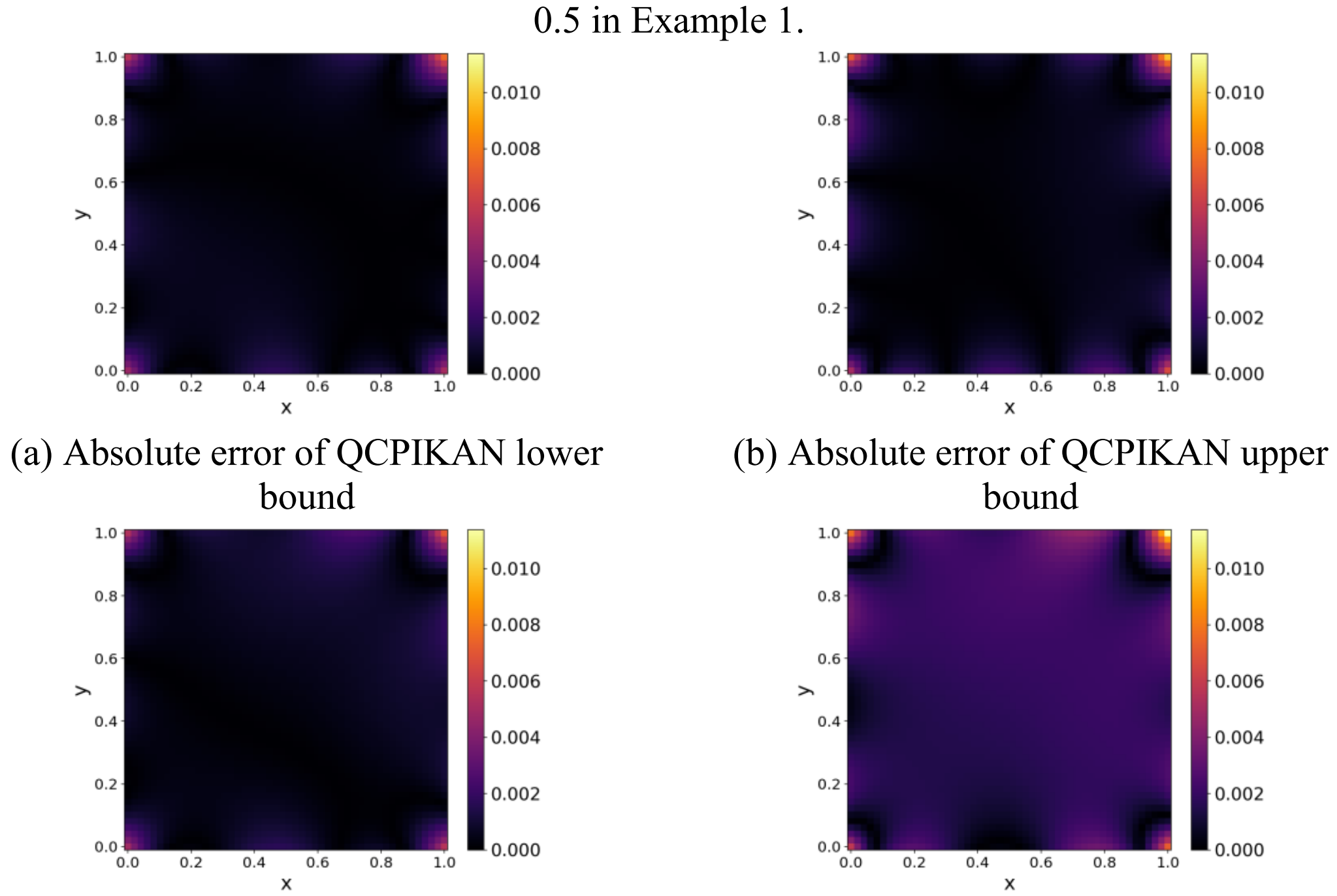

(a) Absolute error of QCPIKAN lower bound

(b) Absolute error of QCPIKAN upper bound

(c) Absolute error of PIKAN lower bound

(d) Absolute error of PIKAN upper bound

Fig. 5. Absolute errors of the lower- and upper-endpoint solutions in Example 1 at $\alpha$ = 0.50.

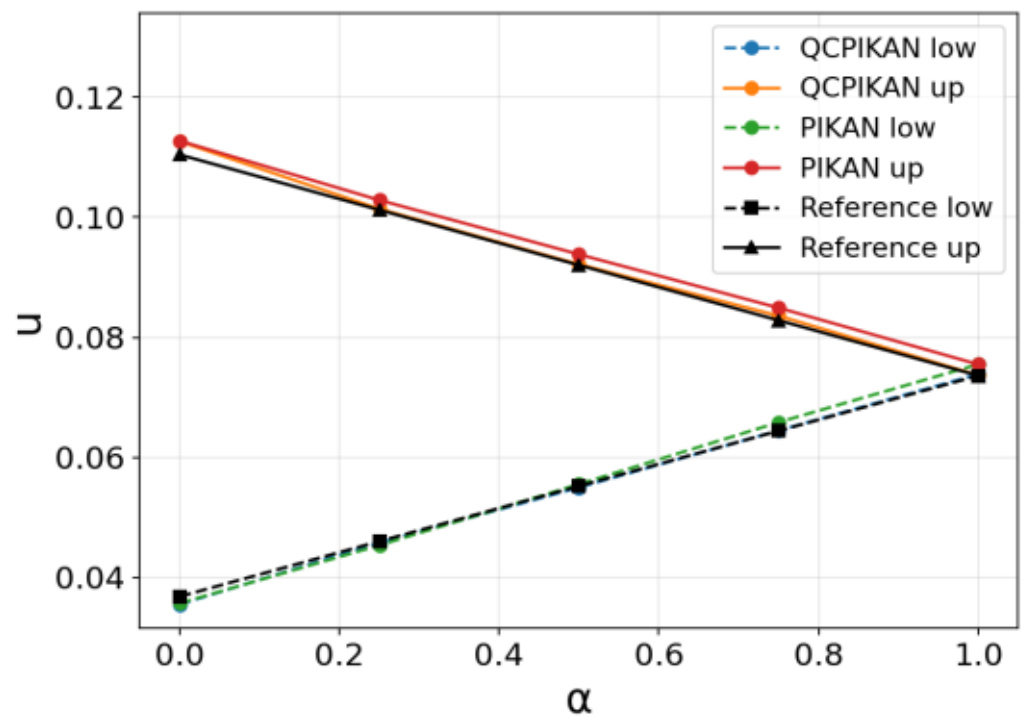

(a) Variation of the predicted and reference endpoints with $\alpha$ at the central location

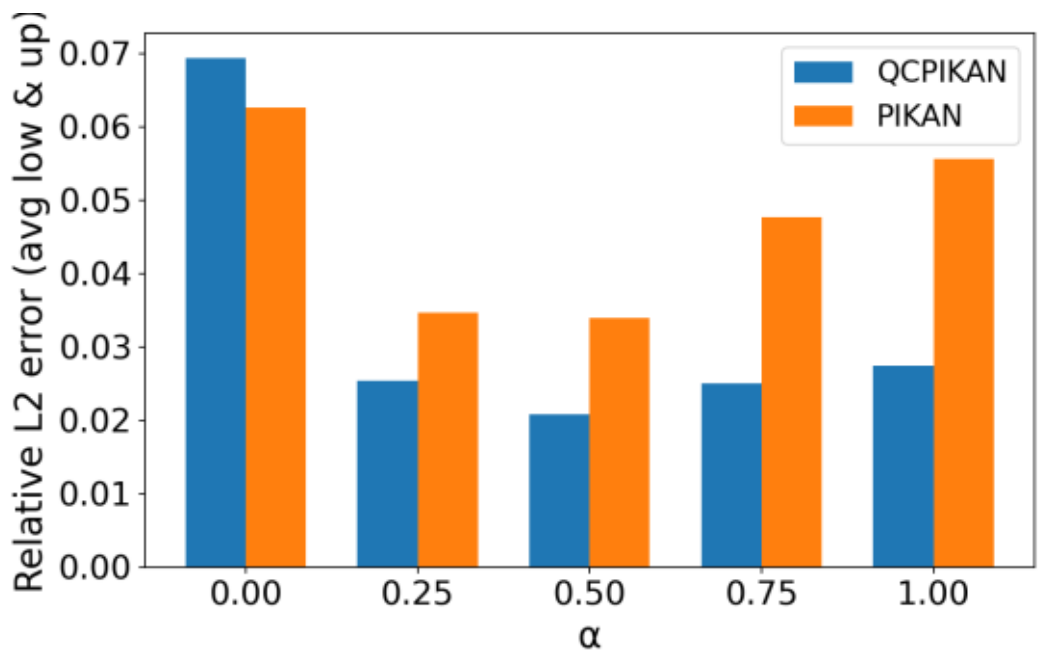

(b) Mean relative $L_2$ error

Fig. 6. Prediction errors and comparison with the reference solution at different $\alpha$ levels in Example 1.

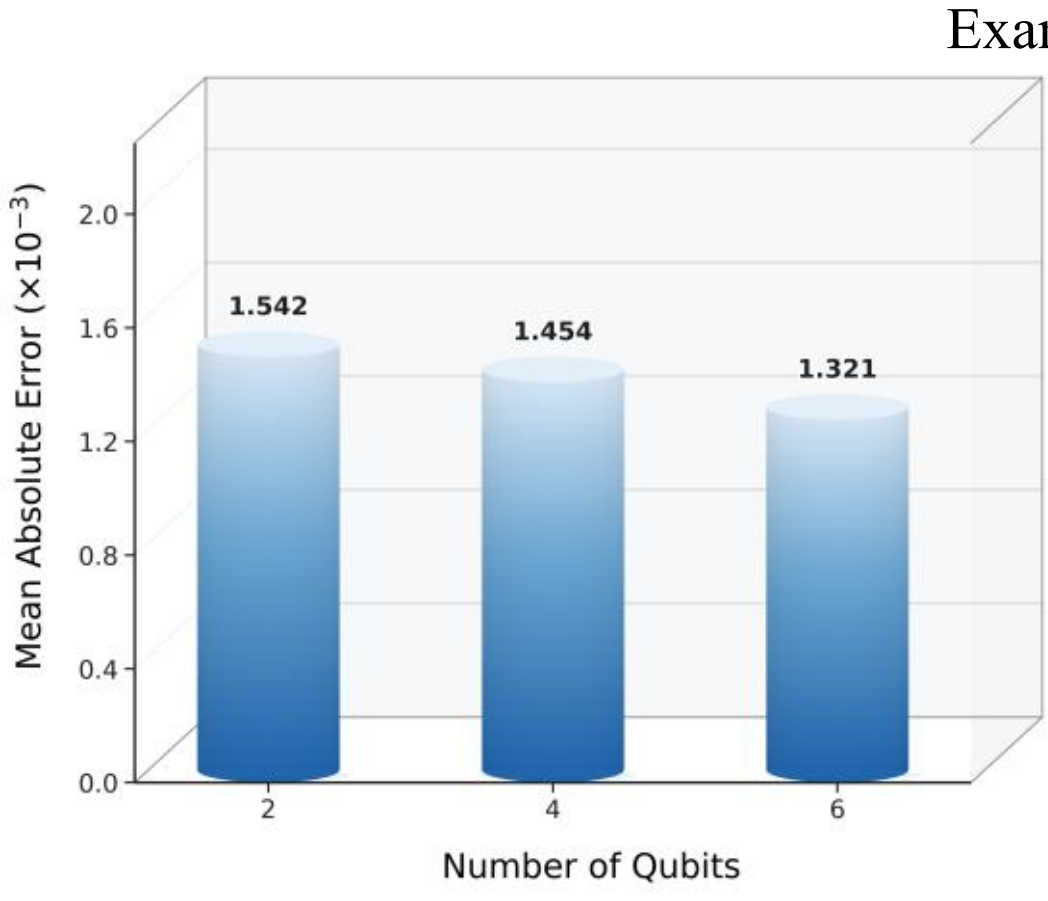

(a) Fixed quantum-circuit depth of one layer

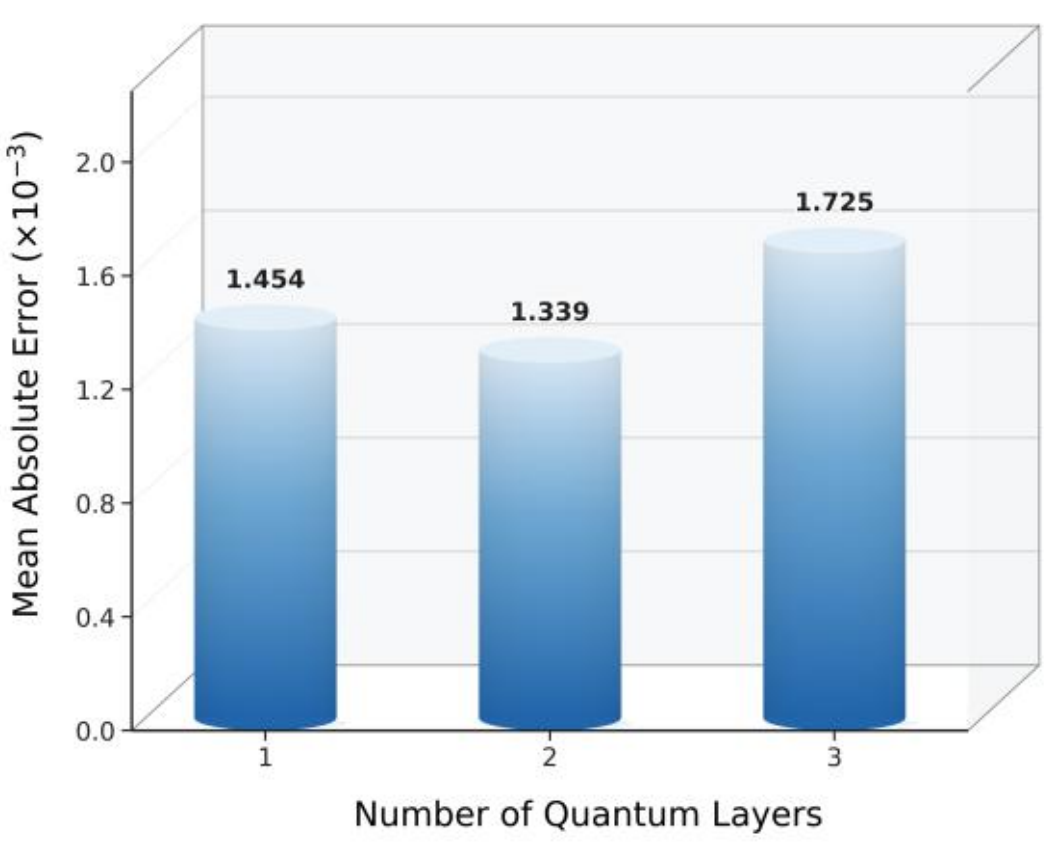

(b) Fixed number of four qubits

Fig. 7. Effects of the number of qubits and parameterized quantum-circuit layers on the mean absolute error of QCPIKAN in Example 1.

3.2 Example 2: Parabolic Fuzzy Heat-Conduction Equation

Consider the following one-dimensional fuzzy heat-conduction problem defined over the space-time domain $(t,x)\in(0,1]\times(0,1)$:

$$\begin{cases}\dfrac{\partial\tilde{u}(t,x)}{\partial t}=\tilde{a}\dfrac{\partial^2\tilde{u}(t,x)}{\partial x^2},(t,x)\in(0,1]\times(0,1),\\ \tilde{u}(0,x)=\sin(\pi x),x\in[0,1],\\ \tilde{u}(t,0)=\tilde{u}(t,1)=0,t\in[0,1],\end{cases}\tag{53}$$

where the diffusion coefficient is the triangular fuzzy number $\tilde{a}=(0.5,1.0,1.5)$, while the initial and boundary conditions are deterministic. In this example, uncertainty in the diffusion coefficient directly affects the decay rate of the solution, making the problem suitable for examining the model's ability to represent the temporal propagation of fuzzy parameters.

At membership level $\alpha\in[0,1]$, the $\alpha$ -cut of the fuzzy diffusion coefficient is

$$[\tilde{a}]_\alpha=\left[a_\alpha^L,a_\alpha^U\right]=[0.5+0.5\alpha,1.5-0.5\alpha].\tag{54}$$

For the initial condition prescribed in Eq. (53), the deterministic heat-equation solution decays faster as the diffusion coefficient increases. Therefore, the lower endpoint of the solution corresponds to $a_\alpha^U$ and the upper endpoint corresponds to $a_\alpha^L$, rather than directly pairing the lower endpoint of the parameter with the lower endpoint of the solution. This yields

$$
\begin{cases}
\dfrac{\partial u_\alpha^L}{\partial t} = a_\alpha^U \dfrac{\partial^2 u_\alpha^L}{\partial x^2}, (t,x) \in (0,1] \times (0,1), \\
\dfrac{\partial u_\alpha^U}{\partial t} = a_\alpha^L \dfrac{\partial^2 u_\alpha^U}{\partial x^2}, (t,x) \in (0,1] \times (0,1), \\
u_\alpha^L(0,x) = u_\alpha^U(0,x) = \sin(\pi x), x \in [0,1], \\
u_\alpha^L(t,0) = u_\alpha^L(t,1) = 0, t \in [0,1], \\
u_\alpha^U(t,0) = u_\alpha^U(t,1) = 0, t \in [0,1].
\end{cases} \tag{55}
$$

Both QCPIKAN and PIKAN take $(t,x,\alpha)$ as input and simultaneously approximate the lower- and upper-endpoint functions in Eq. (55). Because the initial function $\sin(\pi x)$ is an eigenfunction of the spatial Laplace operator, the reference solution can be written directly as

$$
\begin{cases}
u_\alpha^L(t,x) = \sin(\pi x)\exp\left[-a_\alpha^U \pi^2 t\right], \\
u_\alpha^U(t,x) = \sin(\pi x)\exp\left[-a_\alpha^L \pi^2 t\right].
\end{cases} \tag{56}
$$

Within the unified loss framework in Eq. (32), this example uses $\lambda_{\text{res}} = \lambda_{\text{IC}} = \lambda_{\text{BC}} = \lambda_{\alpha=1} = 1$, $\lambda_{\text{nest}} = 0.5$, and $\lambda_{\text{order}} = 0.2$. The resulting loss is therefore $L(\boldsymbol{\theta}) = \mathcal{L}_{\text{res}} + \mathcal{L}_{\text{IC}} + \mathcal{L}_{\text{BC}} + 0.5\mathcal{L}_{\text{nest}} + 0.2\mathcal{L}_{\text{order}} + \mathcal{L}_{\alpha=1}$. The two models use identical training points, loss weights, and optimization parameters.

Figs. 8-10 compare the training behavior, endpoint evolution, and fuzzy-structure properties of the two models. After approximately 2000 epochs, QCPIKAN maintains a lower and less fluctuating loss, and the final PIKAN loss is about 1.7 times larger. Both models capture temporal decay and interval contraction as $\alpha$ increases. PIKAN shows more pronounced negative lower-endpoint values, greater upper-endpoint overestimation, and clearer endpoint separation at $\alpha = 1.00$. QCPIKAN reduces these deviations, although neither model strictly satisfies the nonnegativity requirement or the endpoint-coincidence condition at $\alpha = 1$. The interval-width maps further show wider low-$\alpha$ intervals for PIKAN and local negative widths at high $\alpha$ for both models. QCPIKAN exhibits some interior inter-level nesting violations, whereas such violations are more pronounced for PIKAN near the temporal and spatial boundaries.

Relative to the reference solution, Fig. 11 shows that the largest errors occur near the spatial center at early times and decay thereafter. The PIKAN error regions are broader for both endpoints, particularly for the upper endpoint, whereas QCPIKAN has the lower overall error at $\alpha = 0.50$. In Fig. 12(a), the PIKAN mean relative $L_2$ errors are approximately 2.0, 1.7, and 1.3 times the QCPIKAN errors at $\alpha = 0$, 0.25, and 0.50, respectively; the two models are comparable at $\alpha = 0.75$ and 1.00. Both models track the reference endpoint contraction, but QCPIKAN is more accurate at low $\alpha$, while small deviations from endpoint coincidence remain for both models at $\alpha = 1$.

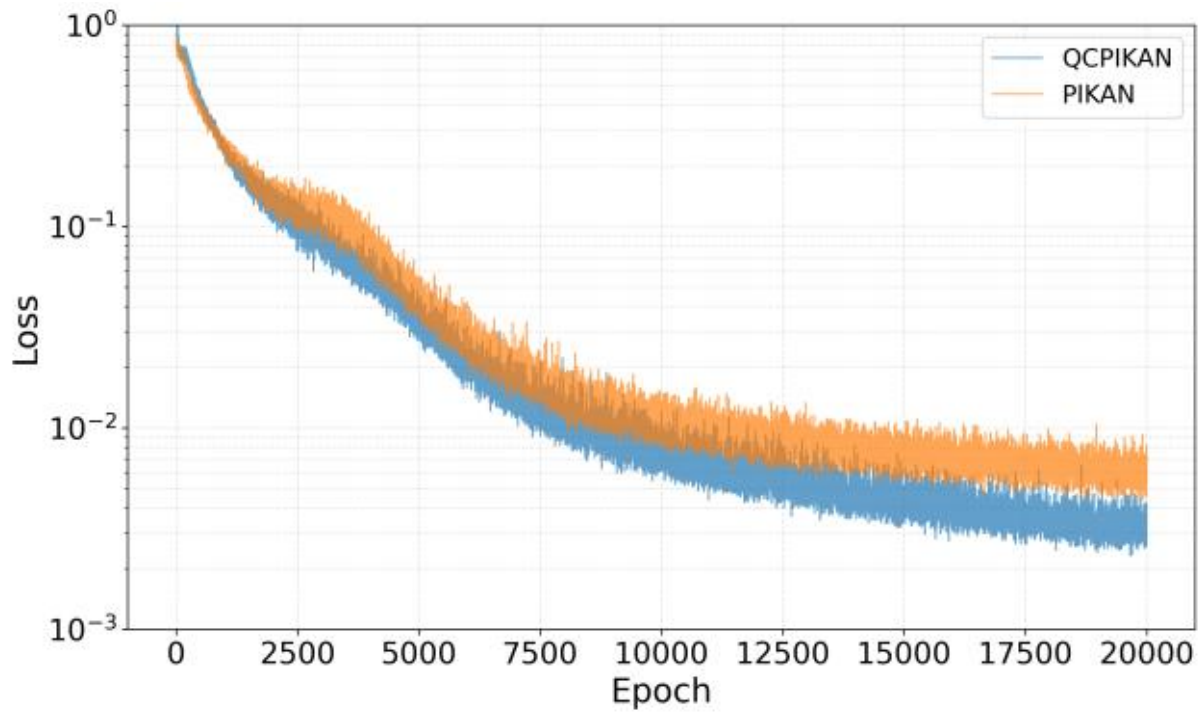

Fig. 8. Comparison of training losses in Example 2.

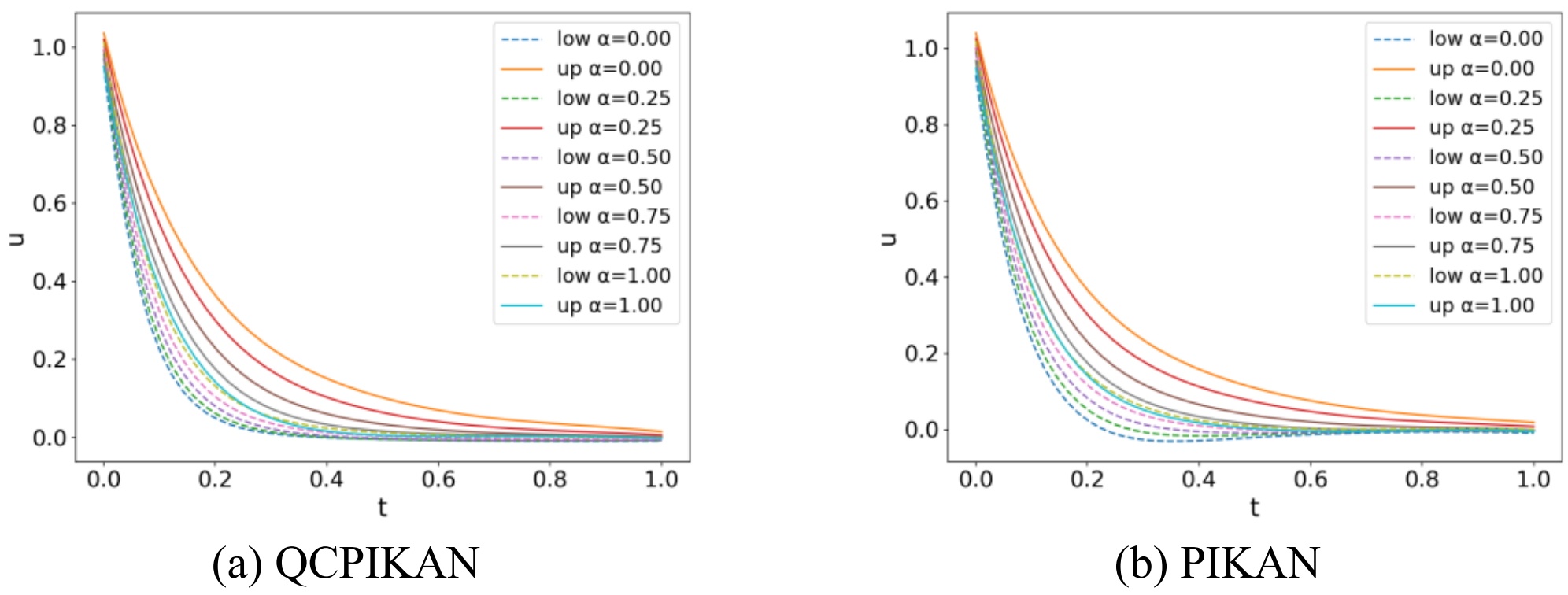

(a) QCPIKAN (b) PIKAN

Fig. 9. Temporal evolution of the lower and upper solution endpoints at $x \approx 0.49$ for different $\alpha$ levels in Example 2.

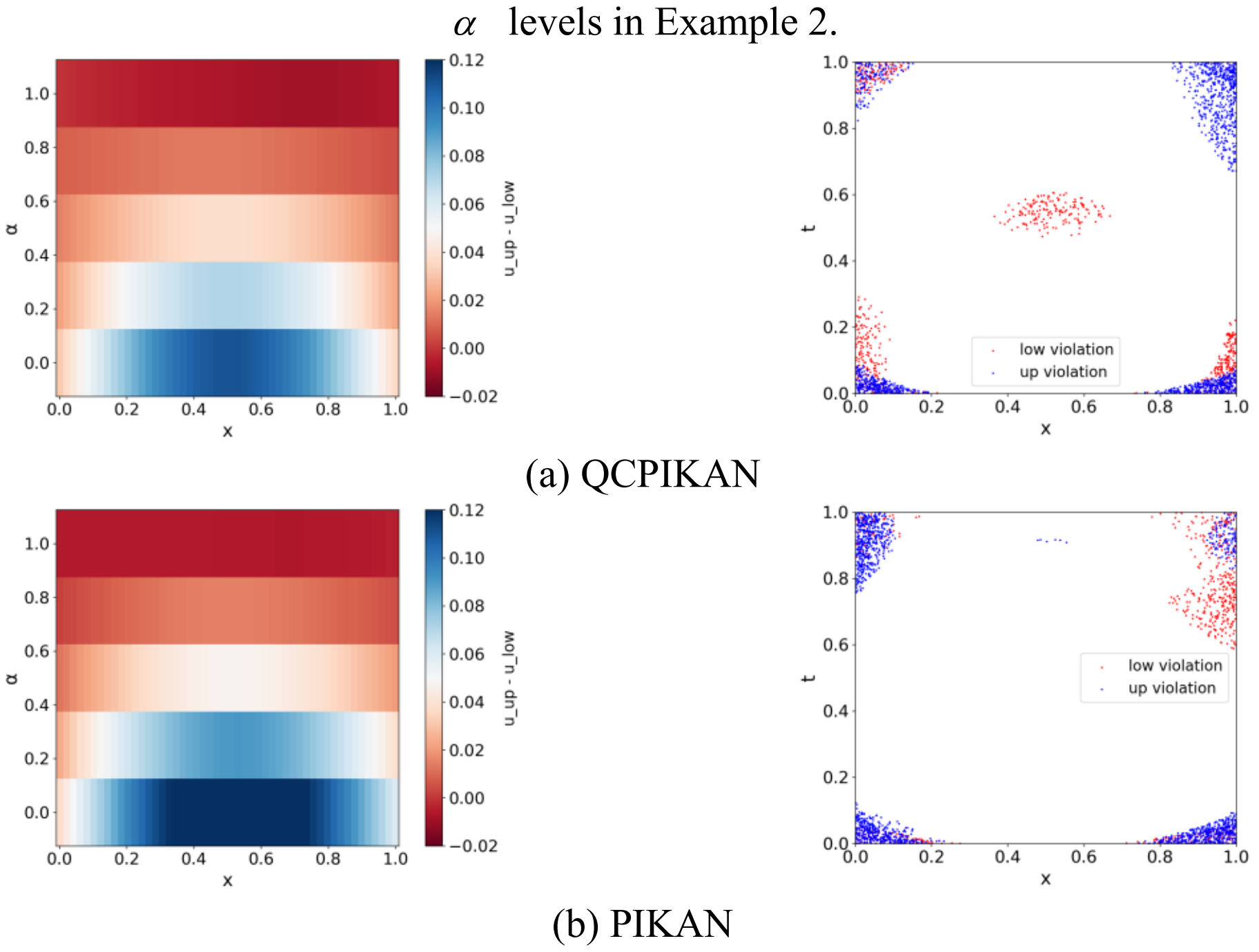

(a) QCPIKAN

(b) PIKAN

Fig. 10. Spatial interval widths at different $\alpha$ levels and inter-level nesting violations along the $\alpha$ direction at $t \approx 0.49$ in Example 2.

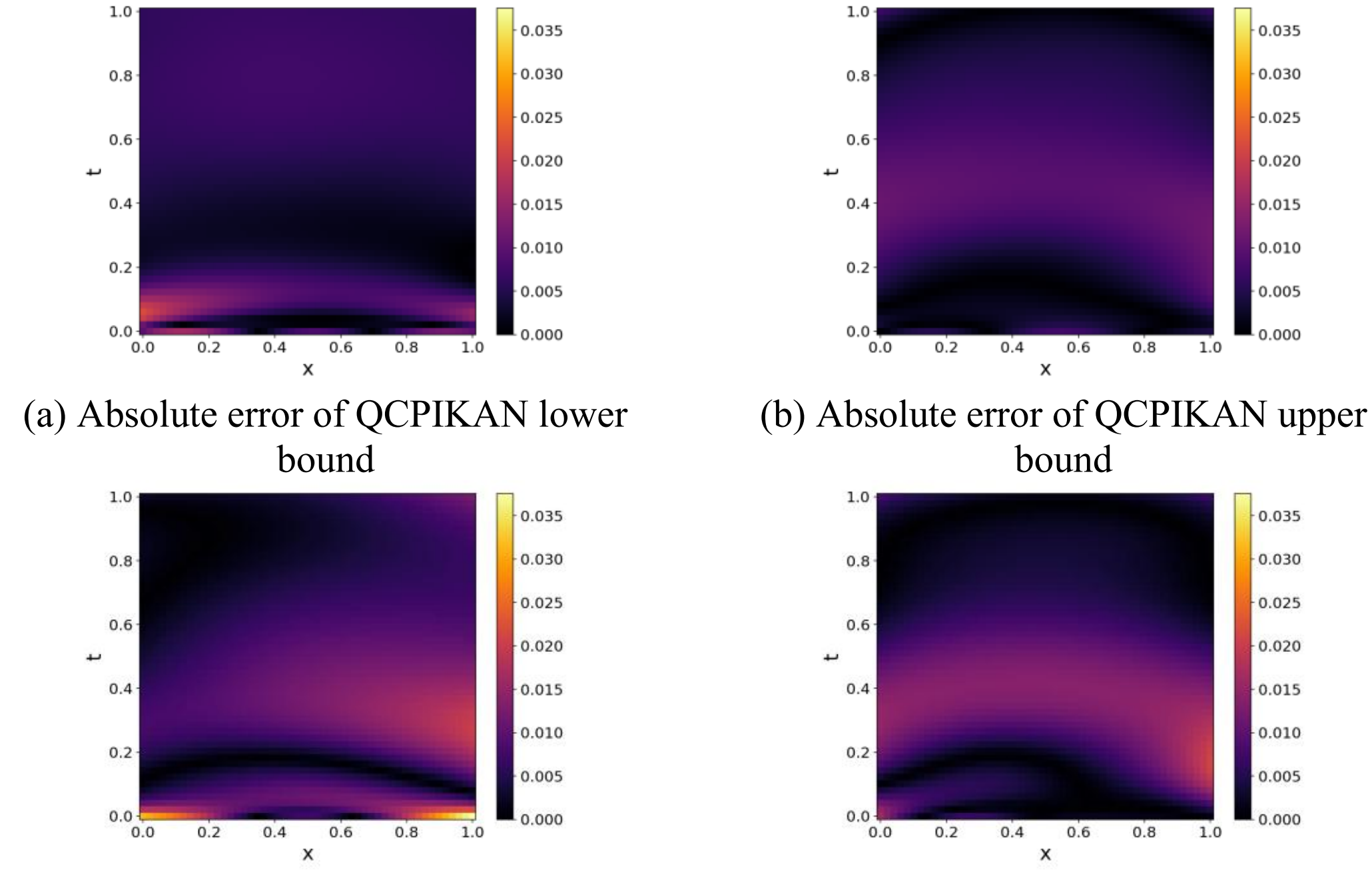

(a) Absolute error of QCPIKAN lower bound

(b) Absolute error of QCPIKAN upper bound

(c) Absolute error of PIKAN lower bound (d) Absolute error of PIKAN upper bound

Fig. 11. Absolute errors of the lower- and upper-endpoint solutions in Example 2 at $\alpha = 0.50$.

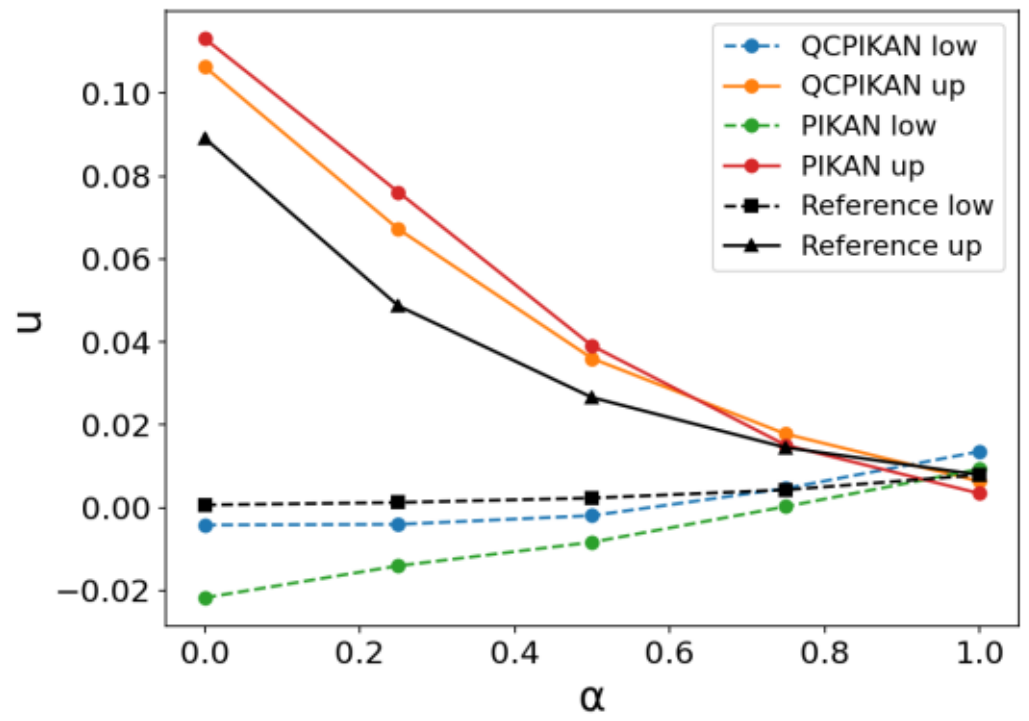

(a) Variation of the predicted and reference endpoints with $\alpha$ at a fixed spatiotemporal location

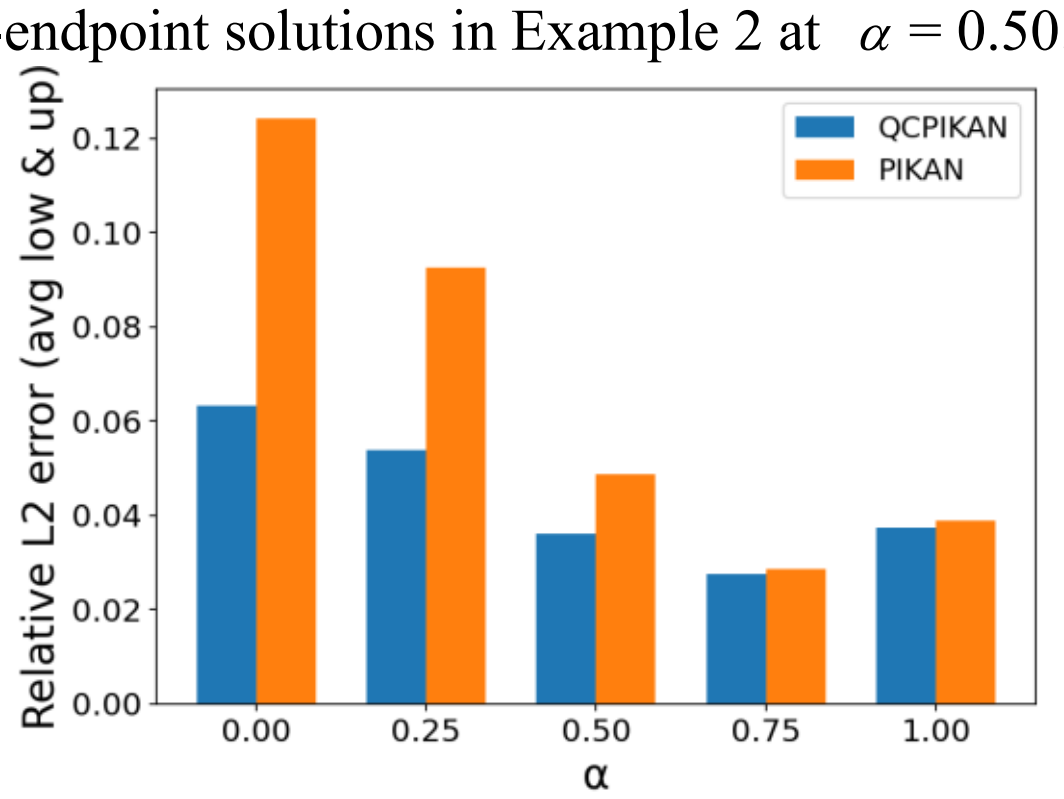

(b) Mean relative $L_2$ error

Fig. 12. Prediction errors and comparison with the reference solution at different $\alpha$ levels in Example 2.

### 3.3 Example 3: Nonlinear Parabolic Fisher-KPP Equation

To further examine the ability of the models to describe the combined effects of diffusion and nonlinear reaction, consider the following Fisher-KPP-type fuzzy reaction-diffusion equation:

$$\begin{cases} \dfrac{\partial \tilde{u}(t,x)}{\partial t} = \tilde{a}\dfrac{\partial^2 \tilde{u}(t,x)}{\partial x^2} + \tilde{b}\,\tilde{u}(t,x)\left[1-\tilde{u}(t,x)\right], (t,x)\in(0,1]\times(0,1), \\ \tilde{u}(0,x) = \sin(\pi x), x\in[0,1], \\ \tilde{u}(t,0) = \tilde{u}(t,1) = 0, t\in[0,1], \end{cases} \tag{57}$$

where $\tilde{a}$ and $\tilde{b}$ are the fuzzy diffusion and reaction coefficients, respectively, both represented by the triangular fuzzy number $(0.5, 1.0, 1.5)$. At membership level $\alpha \in [0,1]$, their $\alpha$-cuts are

$$[a]_\alpha = \left[a_\alpha^L, a_\alpha^U\right] = [0.5+0.5\alpha, 1.5-0.5\alpha], \left[\tilde{b}\right]_\alpha = \left[b_\alpha^L, b_\alpha^U\right] = [0.5+0.5\alpha, 1.5-0.5\alpha]. \tag{58}$$

Under the prescribed initial-boundary conditions, increasing the diffusion coefficient accelerates spatial dissipation, whereas increasing the reaction coefficient strengthens nonlinear growth. Accordingly, the lower endpoint of the solution in the implementation uses the parameter combination $\left(a_\alpha^U, b_\alpha^L\right)$, while the upper endpoint uses $\left(a_\alpha^L, b_\alpha^U\right)$. The corresponding endpoint system is

$$\begin{cases} \dfrac{\partial u_\alpha^L}{\partial t} = a_\alpha^U \dfrac{\partial^2 u_\alpha^L}{\partial x^2} + b_\alpha^L u_\alpha^L\left(1-u_\alpha^L\right), \\ \dfrac{\partial u_\alpha^U}{\partial t} = a_\alpha^L \dfrac{\partial^2 u_\alpha^U}{\partial x^2} + b_\alpha^U u_\alpha^U\left(1-u_\alpha^U\right), \\ u_\alpha^L(0,x) = u_\alpha^U(0,x) = \sin(\pi x), \\ u_\alpha^L(t,0) = u_\alpha^L(t,1) = 0, \\ u_\alpha^U(t,0) = u_\alpha^U(t,1) = 0. \end{cases} \tag{59}$$

Because Eq. (59) contains a nonlinear reaction term, an IMEX Crank-Nicolson finite-difference method is used to construct the numerical reference solution. The diffusion term is discretized using the Crank-Nicolson scheme, the reaction term is treated explicitly, and the resulting tridiagonal linear system is solved using the Thomas algorithm. The reference solution is computed using 200 spatial subintervals and 1000 temporal subintervals and is interpolated onto the model-prediction grid for error evaluation.

Within the unified loss framework in Eq. (32), this example uses $\lambda_{\text{res}} = \lambda_{\text{IC}} = \lambda_{\text{BC}} = \lambda_{\alpha=1} = 1$, $\lambda_{\text{nest}} = 0.1$, and $\lambda_{\text{order}} = 0.2$. The resulting loss is therefore $L(\boldsymbol{\theta}) = \mathcal{L}_{\text{res}} + \mathcal{L}_{\text{IC}} + \mathcal{L}_{\text{BC}} + 0.1\mathcal{L}_{\text{nest}} + 0.2\mathcal{L}_{\text{order}} + \mathcal{L}_{\alpha=1}$. The two models use identical training points, loss weights, and optimization parameters.

Figs. 13-15 compare the training behavior, endpoint evolution, and fuzzy-structure properties of QCPIKAN and PIKAN. QCPIKAN maintains the lower loss throughout most of training, while PIKAN drops markedly at approximately 11000-12000 epochs but still ends with a loss about 2.5 times larger. Both models reproduce temporal decay and interval contraction. However, some PIKAN lower-endpoint curves approach or cross the upper endpoint at later times, whereas the QCPIKAN endpoint separation is more stable. PIKAN also produces wider intervals at low $\alpha$, and both models show local negative widths at high $\alpha$. Despite its lower prediction error, QCPIKAN has more widely distributed inter-level nesting violations, including a clear interior cluster; such violations for PIKAN are concentrated mainly near the boundaries and in the right-hand region. QCPIKAN therefore shows no advantage over PIKAN in maintaining inter-level nesting along the $\alpha$ direction in this example.

Relative to the reference solution, Fig. 16 shows smaller and more localized QCPIKAN errors, whereas PIKAN forms broader error bands for both endpoints, especially for the upper endpoint at low to intermediate times. At $\alpha$ = 0, 0.25, 0.50, 0.75, and 1.00, Fig. 17(a) gives PIKAN-to-QCPIKAN mean relative $L_2$ error ratios of approximately 2.4, 2.7, 2.7, 1.5, and 1.1, respectively. Fig. 17(b) further shows that both models capture endpoint contraction; the QCPIKAN lower endpoint is generally closer to the reference curve, while both models overestimate the upper endpoint, more noticeably for PIKAN at low to intermediate $\alpha$. Neither model fully satisfies the endpoint-coincidence condition at $\alpha = 1$.

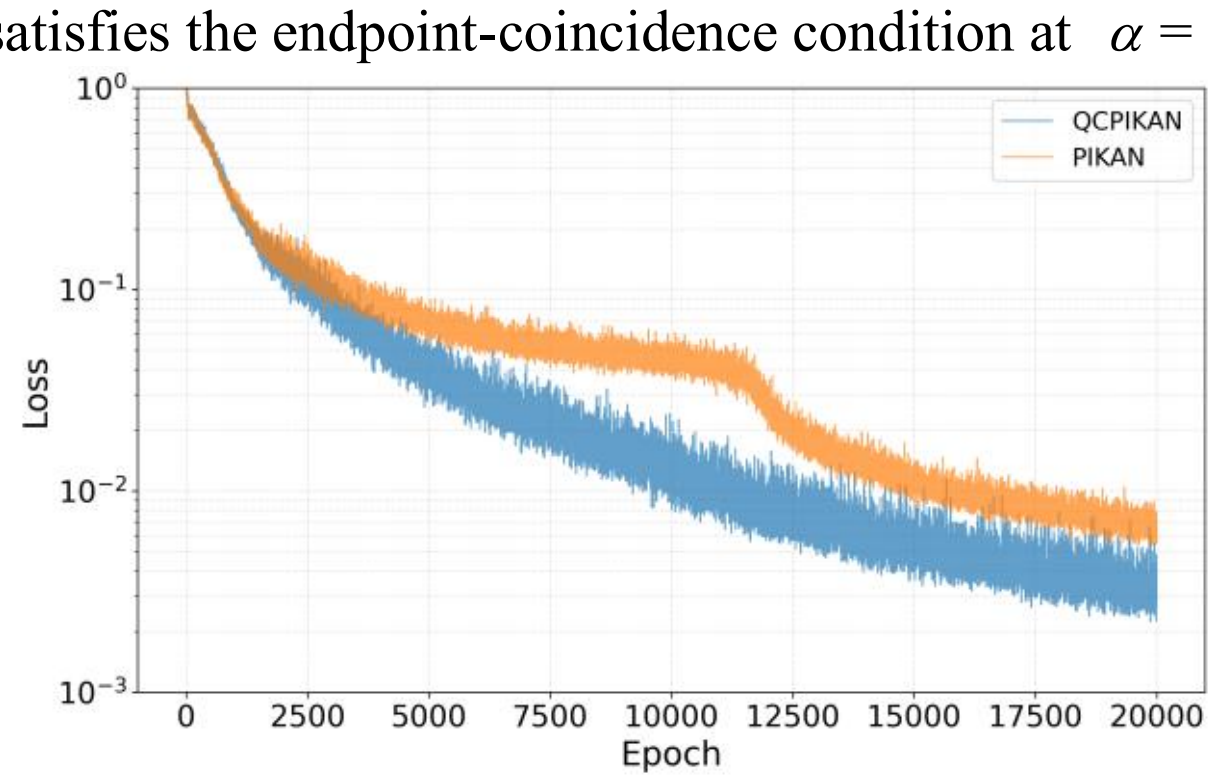

Fig. 13. Comparison of training losses in Example 3.

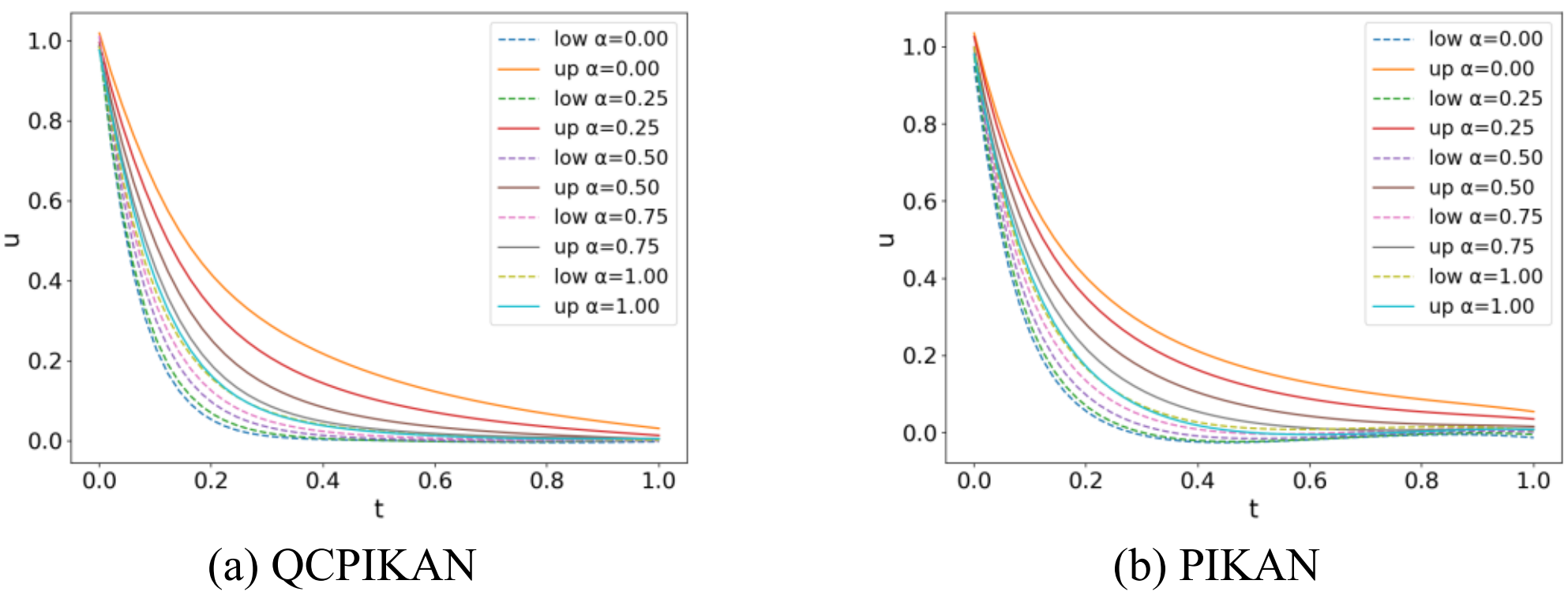

(a) QCPIKAN (b) PIKAN

Fig. 14. Temporal evolution of the lower and upper solution endpoints at $x \approx 0.49$ for different $\alpha$ levels in Example 3.

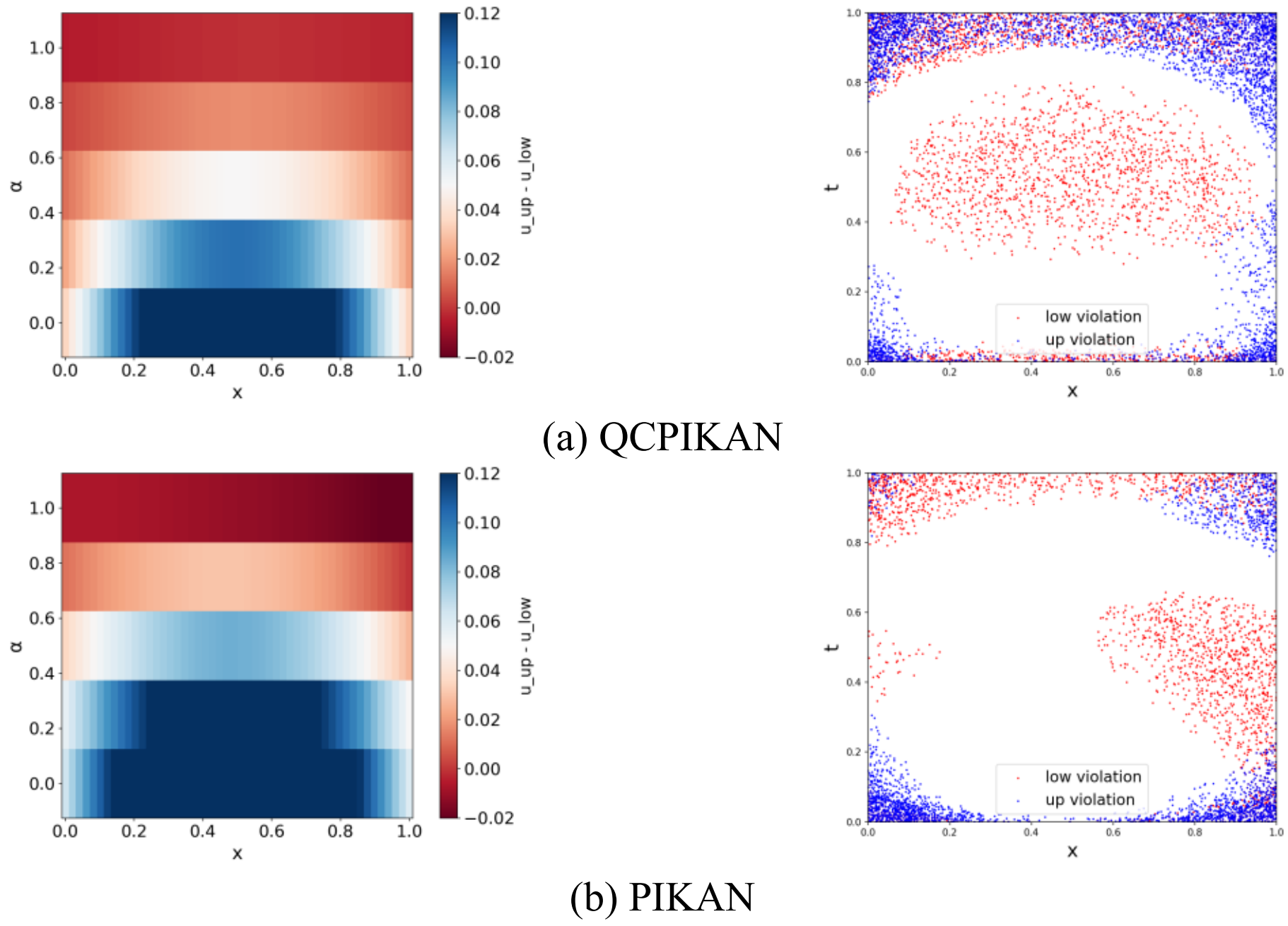

(a) QCPIKAN

(b) PIKAN

Fig. 15. Spatial interval widths at different $\alpha$ levels and inter-level nesting violations along the $\alpha$ direction at $t \approx 0.49$ in Example 3.

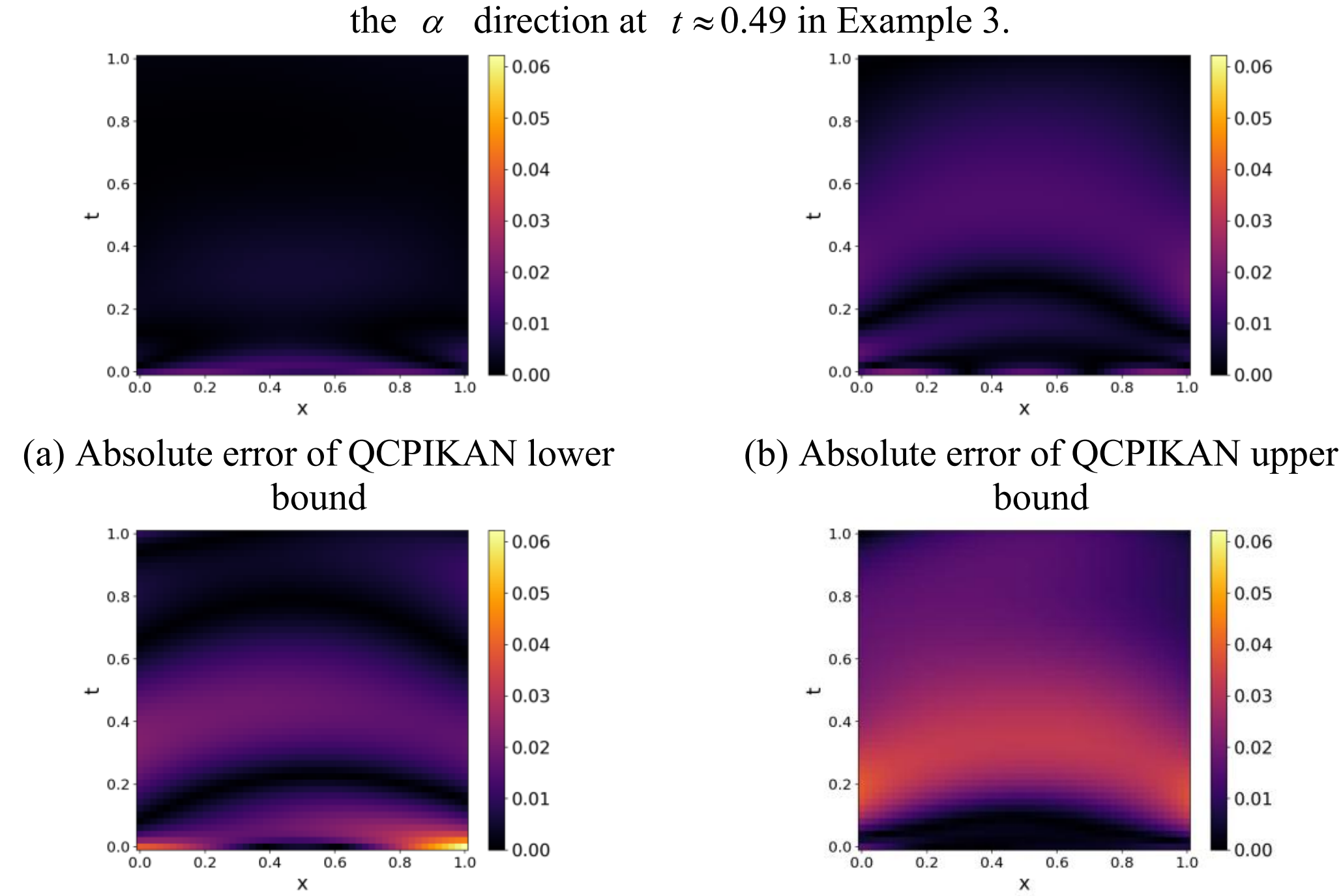

(a) Absolute error of QCPIKAN lower bound

(b) Absolute error of QCPIKAN upper bound

(c) Absolute error of PIKAN lower bound

(d) Absolute error of PIKAN upper bound

Fig. 16. Absolute errors of the lower- and upper-endpoint solutions in Example 3 at $\alpha = 0.50$.

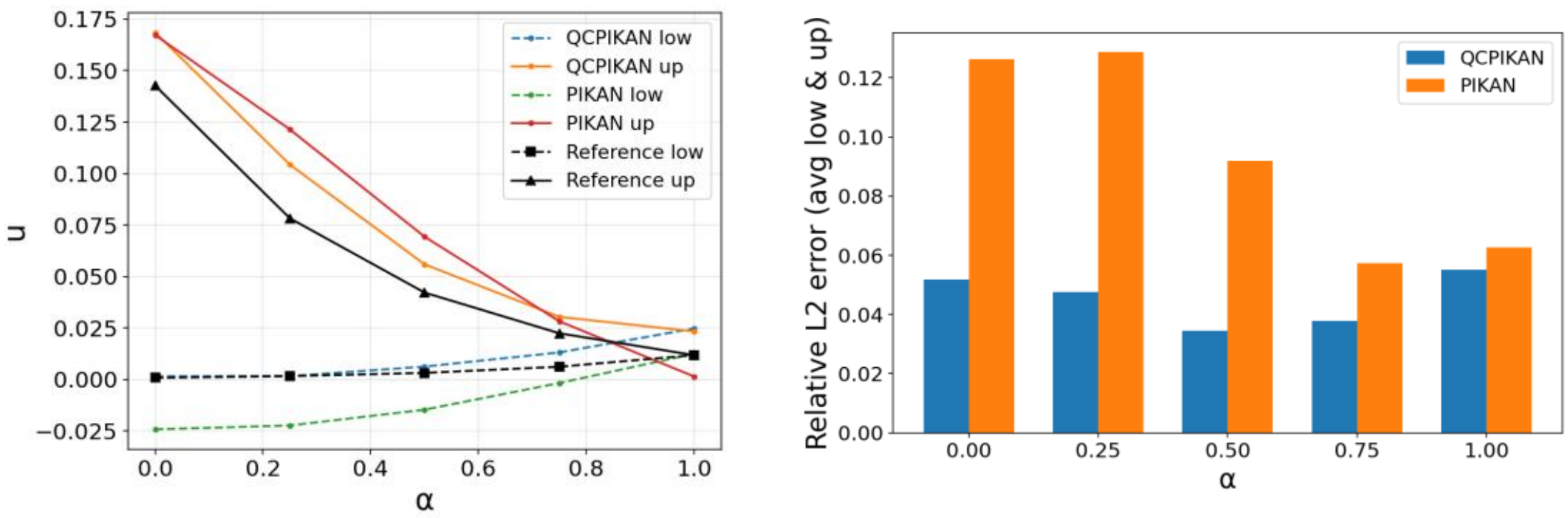

(a) Variation of the predicted and reference endpoints with $\alpha$ at a fixed spatiotemporal location

(b) Mean relative $L_2$ error

Fig. 17. Prediction errors and comparison with the reference solution at different $\alpha$ levels in Example 3.

3.4 Example 4: Hyperbolic Fuzzy Convection Equation

To evaluate the ability of the models to represent a moving wavefront and its positional uncertainty in a propagation-dominated problem, consider the following one-dimensional linear convection equation with a fuzzy convection velocity:

$$\begin{cases} \frac{\partial \tilde{u}}{\partial t} + \tilde{c}\frac{\partial \tilde{u}}{\partial x} = 0, (t,x) \in (0,0.5] \times (0,1), \\ \tilde{u}(0,x) = u_0(x), x \in [0,1], \\ \tilde{u}(t,0) = u_0(-\tilde{c}t), t \in (0,0.5], \end{cases} \tag{60}$$

where the initial function is $u_0(x) = \frac{1}{1+\exp[-(x-0.5)/0.08]}$ and the fuzzy convection velocity is the triangular fuzzy number $\tilde{c} = (0.8,1.0,1.2)$. Because its values are always positive, characteristics enter the computational domain only through the left boundary. Therefore, Eq. (60) requires only a left inflow boundary condition.

At a prescribed membership level $\alpha$, the $\alpha$-cut of the convection velocity is

$$[\tilde{c}]_\alpha = [c_\alpha^L, c_\alpha^U] = [0.8+0.2\alpha, 1.2-0.2\alpha], \alpha \in [0,1]. \tag{61}$$

The initial function $u_0(x)$ is monotonically increasing with respect to $x$, whereas $u_0(x-ct)$, at fixed $(t,x)$, decreases as $c$ increases. Therefore, the lower endpoint of the solution corresponds to $c_\alpha^U$ and the upper endpoint corresponds to $c_\alpha^L$. The corresponding endpoint problem is

$$\begin{cases} \frac{\partial u_\alpha^L}{\partial t} + c_\alpha^U \frac{\partial u_\alpha^L}{\partial x} = 0, \\ \frac{\partial u_\alpha^U}{\partial t} + c_\alpha^L \frac{\partial u_\alpha^U}{\partial x} = 0, \\ u_\alpha^L(0,x) = u_\alpha^U(0,x) = u_0(x), \\ u_\alpha^L(t,0) = u_0(-c_\alpha^U t), \\ u_\alpha^U(t,0) = u_0(-c_\alpha^L t). \end{cases} \tag{62}$$

According to the method of characteristics, the analytical endpoint solutions of Eq. (62) are

$$u_\alpha^L(t,x) = u_0(x - c_\alpha^U t), u_\alpha^U(t,x) = u_0(x - c_\alpha^L t). \tag{63}$$

Eq. (63) is evaluated on a 201 × 201 space-time grid, and reference solutions are constructed at $\alpha$ = 0, 0.25, 0.50, 0.75, and 1.00.

The model takes $(t,x,\alpha)$ as input and simultaneously outputs $u_\alpha^L$ and $u_\alpha^U$. According to Eq. (32), the loss weights used in this example are $\lambda_{\text{res}} = 5$, $\lambda_{\text{IC}} = 10$, $\lambda_{\text{BC}} = 5$, $\lambda_{\text{nest}} = 0.5$, $\lambda_{\text{order}} = 0.15$, and $\lambda_{\alpha=1} = 0.15$. The two models use identical training points, loss weights, and optimization parameters.

Figs. 18-20 compare the training behavior, wavefront propagation, and fuzzy-structure properties of the two models. Their losses decrease rapidly at first and then more gradually, reaching the same order of magnitude. Both models reproduce the right-moving inclined wavefront, and the lower-endpoint front leads the upper-endpoint front because it corresponds to the larger convection velocity. The interval width is concentrated along the moving front and is nearly zero in the constant states on either side, showing that velocity uncertainty is expressed

mainly as wavefront-position uncertainty. Inter-level nesting violations also cluster near the wavefront and the boundaries. Because PIKAN has fewer upper-endpoint inter-level nesting violations, QCPIKAN does not show a consistent advantage in the fuzzy-structure metrics.

To quantitatively evaluate the wavefront position, the intersection of a predicted profile with $u = 0.50$ is defined as the wavefront position, and its coordinate is determined by linear interpolation between adjacent grid points. Let $\mathcal{V}_\alpha$ denote the endpoint-time combinations for which valid intersections exist in both the predicted and reference solutions. The mean wavefront-position error is defined as

$$E_{\mathrm{wf}}(\alpha)=\frac{1}{|\mathcal{V}_\alpha|}\sum_{(b,t_j)\in\mathcal{V}_\alpha}\left|\hat{x}_{\mathrm{wf},\alpha}^{b}(t_j)-x_{\mathrm{wf},\alpha,\mathrm{ref}}^{b}(t_j)\right|, b\in\{L,U\}, t_j\in\{0,0.25,0.5\}. \tag{64}$$

At the five reference levels $\alpha = 0$, 0.25, 0.50, 0.75, and 1.00, Fig. 21(a) shows that both models closely reproduce the S-shaped reference wavefront at $t = 0.24$; the endpoint gap narrows as α increases and is nearly closed at $\alpha = 1.00$, with the remaining deviations confined mainly to the steep transition. Fig. 21(b) shows lower QCPIKAN wavefront-position errors at every level. The PIKAN errors are approximately 1.13-3.24 times the QCPIKAN errors and 1.77 times larger on average. The largest difference occurs at $\alpha = 0.50$ (approximately 3.24 times), whereas the difference is smaller near $\alpha = 0$ and 1.00. QCPIKAN therefore captures the wavefront position and propagation phase more accurately in this example.

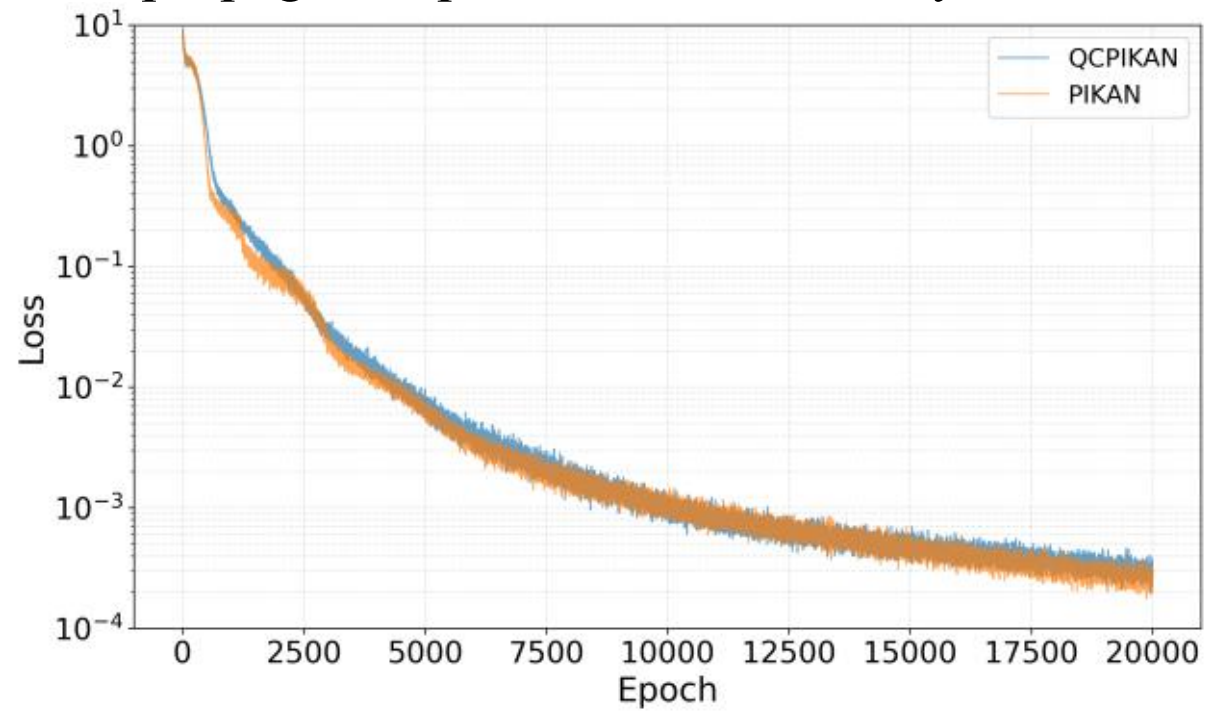

Fig. 18. Comparison of training losses in Example 4.

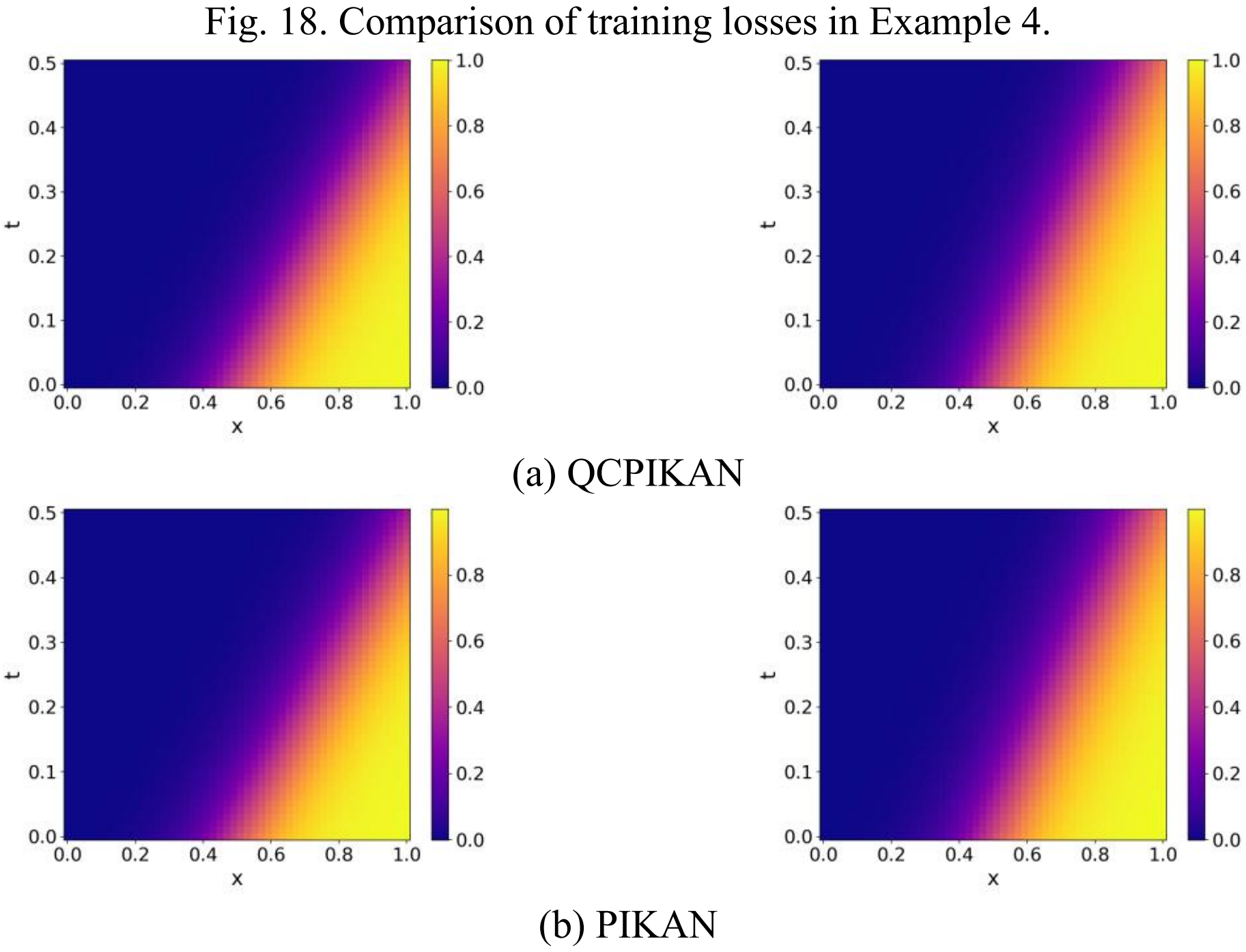

Fig. 19. Space-time distributions of the lower- and upper-endpoint solutions in Example 4 at $\alpha = 0.50$.

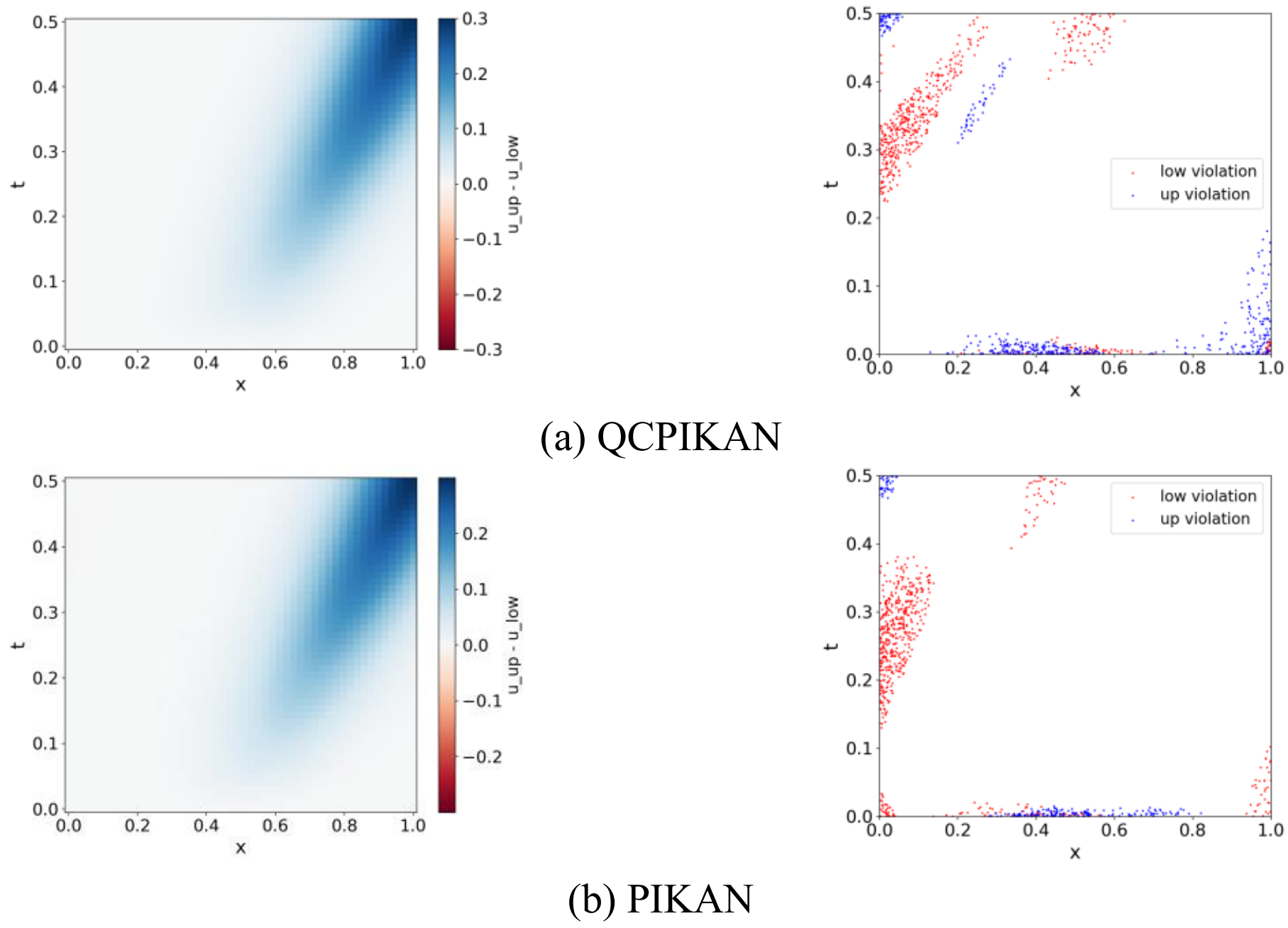

(a) QCPIKAN

(b) PIKAN

Fig. 20. Space-time interval width and inter-level nesting violations along the $\alpha$ direction at $\alpha = 0.5$ in Example 4.

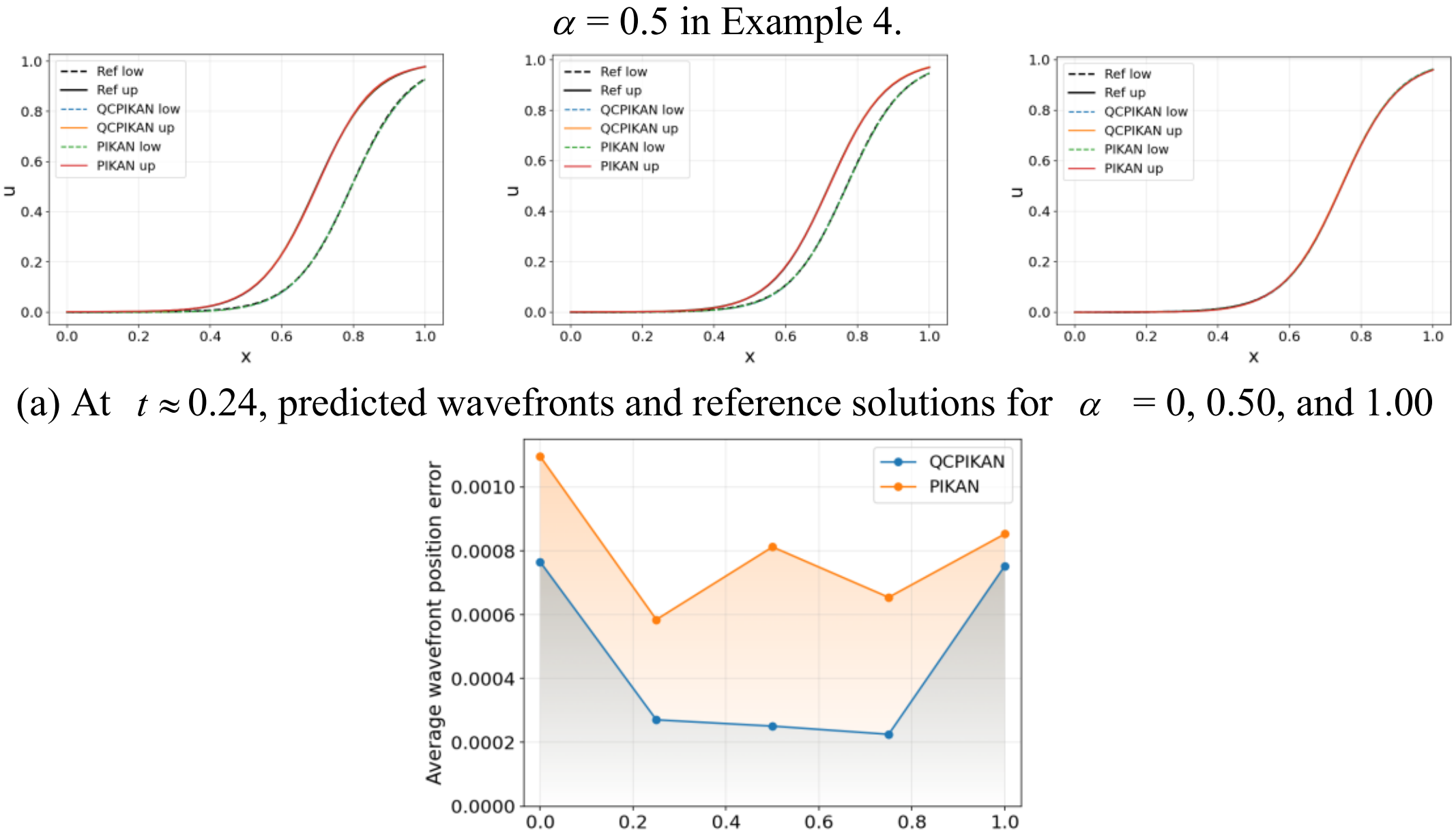

(a) At $t \approx 0.24$, predicted wavefronts and reference solutions for $\alpha$ = 0, 0.50, and 1.00

(b) Mean wavefront-position errors at different $\alpha$ levels

Fig. 21. Wavefront-profile predictions and mean position errors in Example 4.

## 4 Conclusions

This study constructs a QCPIKAN framework for solving fuzzy partial differential equations represented by $\alpha$ -cuts. ChebyKAN modules and a parameterized quantum circuit are used to jointly represent the dependence of the solution on the spatiotemporal variables and membership level. The lower- and upper-endpoint functions are trained in a unified manner using the governing equations, initial-boundary conditions, endpoint ordering, inter-level nesting, and the endpoint-coincidence loss at $\alpha = 1$. The theoretical analysis shows that the accuracy advantage of QCPIKAN is not automatically guaranteed by the quantum module. Its theoretical endpoint-solution error bound can be strictly smaller than that of PIKAN when the representation gain provided by the hybrid feature space is sufficient to offset the optimization and finite-sampling errors, as well as the fuzzy-structure constraint error arising from soft enforcement.

The numerical experiments cover elliptic, parabolic, and hyperbolic FPDEs. In the fuzzy Poisson example, except at $\alpha = 0$, the mean relative $L_2$ errors of PIKAN are approximately 1.4-2.0 times those of QCPIKAN. At low to intermediate membership levels in the fuzzy heat-conduction example, the corresponding ratios are approximately 1.3-2.0. In the fuzzy reaction-diffusion example, QCPIKAN yields lower errors at all tested levels, with the PIKAN errors being approximately 1.1-2.7 times larger. For the fuzzy convection equation, the wavefront-position errors of PIKAN are approximately 1.13-3.24 times those of QCPIKAN, and the average over the five membership levels is approximately 1.77 times larger.

## 5 Acknowledgments

Prof. Rao acknowledges financial support from the General Program of National Natural Science Foundation of China (NSFC) (No. 52574028), the National Science and Technology Major Project of China (No. 2025ZD1401106), the General Program of Natural Science Foundation of Hubei Province (No. 2026AFB760), and the Xinjiang Uygur Autonomous Region Scientific and Technological Innovation Team Project (No. 2024TSYCTD0018).